\pdfoutput=1

\documentclass[11pt]{article}

\usepackage[final]{acl}

\usepackage{times}
\usepackage{latexsym}
\usepackage[most]{tcolorbox}
\usepackage{xcolor}
\usepackage{float}
\usepackage{flushend}
\usepackage{multirow}

\usepackage[T1]{fontenc}

\usepackage[utf8]{inputenc}

\usepackage{microtype}

\usepackage{inconsolata}

\usepackage{graphicx}

\title{MILE-RefHumEval: A Reference-Free, Multi-Independent LLM Framework for Human-Aligned Evaluation}

\author{
Nalin Srun\textsuperscript{1},
Parisa Rastin\textsuperscript{1},
Guénaël Cabanes\textsuperscript{1},
Lydia Boudjeloud-Assala\textsuperscript{2} \\
\textsuperscript{1}Université de Lorraine, CNRS, LORIA, F-54000 Nancy, France \\ 
\textsuperscript{2}Université de Lorraine, CNRS, LORIA, F-57000 Metz, France \\ 
\{firstname.lastname\}@univ-lorraine.fr
}

\begin{document}
\maketitle
\begin{abstract}
We introduce MILE-RefHumEval, a reference-free framework for evaluating Large Language Models (LLMs) without ground-truth annotations or evaluator coordination. It leverages an ensemble of independently prompted evaluators guided by a human-aligned schema, supporting both discrete and continuous scoring judgement. With task-specific prompts from best candidate selection, summarization and image captioning to dialogue, MILE-RefHumEval provides flexible, interpretable, and scalable assessments. Experiments show it aligns closely with human judgments, outperforms prior methods, and reduces computational overhead, offering an efficient, robust, and human-aligned solution for real-world LLM evaluation.
\end{abstract}


\section{Introduction}

Large language models (LLMs) have transformed NLP, enabling fluent generation, complex reasoning, and domain adaptation. However, evaluating LLMs, especially for tasks involving structured synthesis, factual accuracy, or domain-specific reasoning, remains difficult. 
Common metrics like BLEU~\cite{papineni2002bleu}, ROUGE~\cite{lin2004rouge}, METEOR~\cite{banerjee-lavie-2005-meteor}, and BERTScore~\cite{Zhang*2020BERTScore:} depend on reference outputs and struggle with semantic nuance, dynamic placeholders, and compositional structure.

Recent LLM-as-judge approaches cooperative~\cite{liang2024abseval,xu2023towards}, competitive~\cite{chan2023CHATEVAL,zhao2024auto}, and aggregation-based~\cite{ning2025pico,shu-etal-2024-fusion} have advanced the field but remain sensitive to evaluator bias, inter-model noise, or lack interpretability~\cite{liu-etal-2024-empirical}.

We introduce MILE-RefHumEval, a reference-free, multi-evaluator framework where each LLM assesses candidate responses independently using a shared, human-aligned judgment criteria. This structure eliminates interaction bias and enhances objectivity. We aim to investigate how multiple evaluators independently assess a single role, highlighting the distinct reasoning and judgments that emerge without interaction.

Importantly, we 
designed the framework to support flexible evaluation across a wide range of tasks without the need for excessive time to redesign model architectures to adapt to new tasks. The framework allows effortless adaptation. Users simply modify the evaluation prompt criteria based on the specific task and swap the evaluator type. For example, image-to-text evaluators can be used for image captioning, and text-to-text evaluators can be used for summarization. This enables seamless evaluation across diverse modalities and tasks.  


Our main contributions are: (1) a reference-free, task-agnostic evaluation framework suited for complex, open-ended tasks; (2) an unbiased scoring design using isolated LLM evaluators to prevent cross-influence; 
(3) task-dependent evaluation dimensions that can be flexibly adapted to different benchmarks; (4) evaluator-type interchangeability that enables users to seamlessly switch modalities; and (5) an efficient structure that reduces query load while preserving evaluation quality.

\section{Related Work}
Collaborative frameworks like ABSEval~\citep{liang2024abseval} use role-specialized agents, e.g., commonsense reasoners and code executors, to produce multi-perspective assessments. Others, such as peer-review-style systems~\citep{xu2023towards}, involve iterative critique and revision among agents, emulating human feedback cycles. While these methods enhance interpretability and simulate diverse reasoning, they often suffer from consensus bias, where shared context inflates agreement even on flawed outputs. Debate-style setups assess model quality through structured adversarial exchanges. Auto-Arena~\citep{zhao2024auto} and CHATEVAL~\citep{chan2023CHATEVAL} have models argue under LLM supervision or peer-ranking. JudgeLM~\citep{wang2024judgelm} and DebateSum~\citep{zhang2023debatesum} use third-party judges to rate rhetorical strength. MORE and SAMRE~\citep{bandi2025adversarial} introduce advocate roles and multi-round scoring. These setups better surface qualitative differences but risk dominance bias, where verbosity skews judgment and lowers inter-rater reliability. Ensemble and optimization-driven methods aim for robustness and cost-efficiency. PiCO~\citep{ning2025pico} uses learnable weights to merge rankings from multiple LLMs, while AIME~\citep{patel2024aime} assigns diverse roles to evaluators for adversarial robustness. PoLL~\citep{verga2024replacingjudgesjuriesevaluating} reduces reliance on large models via lightweight mixtures. Other work incorporates voting~\citep{badshah2024reference}, confidence-based cascades~\citep{jung2025trust}, and hybrid metrics~\citep{shu-etal-2024-fusion}, though many require references or are compute-heavy. Domain-specific solutions include recommendation evaluation~\citep{10.1109/TKDE.2024.3392335} and autonomous exam setups~\citep{NEURIPS2023_f64e55d0}. In contrast
, MILE-RefHumEval introduces a novel evaluation paradigm 
both reference-free and interaction-free, relying on multiple independently prompted LLMs to assess a single role without cross-agent influence. This design explicitly avoids consensus or dominance biases introduced by conversational or role-overlapping setups, while still producing human-aligned, interpretable judgments. To our knowledge, it is the first framework to combine evaluator independence with a structured, multi-criteria schema in a fully decentralized manner.

\section{Proposed Framework Design}

\begin{figure}[ht!]
\centering
  \includegraphics[width=\columnwidth]{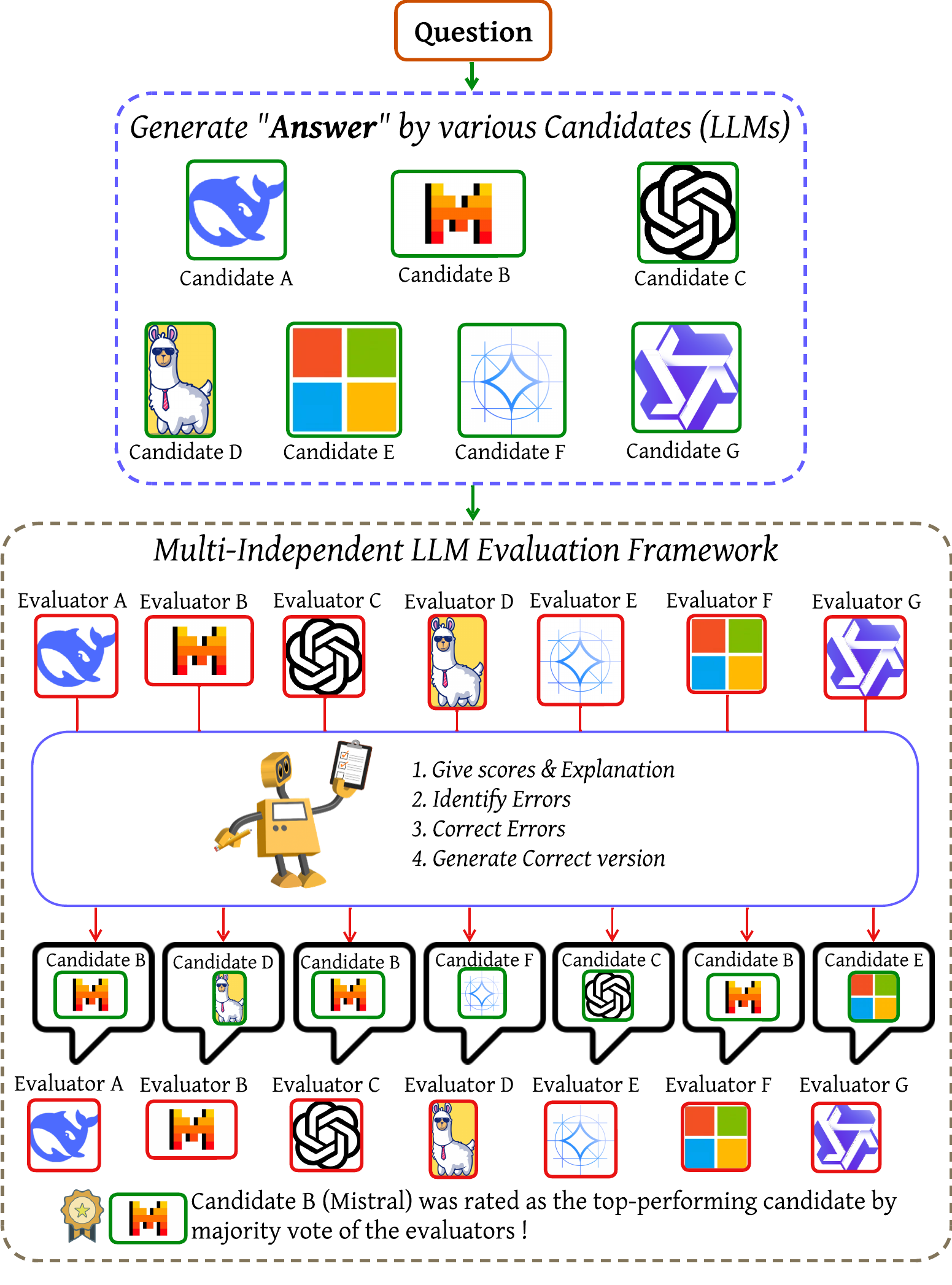}
  \caption{Overall Framework of MILE-RefHumEval.}
  \label{fig:method}
\end{figure}

As illustrated in Figure~\ref{fig:method}, MILE-RefHumEval adopts a reference-free, multi-stage procedure for evaluating LLM-generated responses.
Each candidate output is independently evaluated by a diverse ensemble of evaluators using a unified set of evaluation dimensions designed to comprehensively assess response quality. The specific dimensions are task-dependent and can be flexibly adapted based on the evaluation benchmark.

For example, Figure~\ref{fig:evaluation_criteria_pairwise_comparison} shows that in the candidate comparison task, the evaluation focuses on identifying the best candidate.

\begin{figure}[t]
\centering
\begin{tcolorbox}[
    colframe=black,
    colback=orange!10,
    boxrule=0.4pt,
    arc=3pt,                
    width=\columnwidth,     
    fontupper=\scriptsize,  
    enhanced
]
\textbf{You are Evaluator "A"}, assigned to assess the quality of two candidate answers for the given question. \\[4pt]
--- Question \& Candidate Answers ---
\\ \texttt{\{str(instruction)\}} \\[4pt]
\textbf{[The Start of Question]} \\
\texttt{\{str(input)\}} \\
\textbf{[The End of Question]} \\
\textbf{[The Start of Assistant 1’s Answer]} \\
\texttt{\{response1\}} \\
\textbf{[The End of Assistant 1’s Answer]} \\
\textbf{[The Start of Assistant 2’s Answer]} \\
\texttt{\{response2\}} \\
\textbf{[The End of Assistant 2’s Answer]} \\
---
\\[-2pt]
\textbf{Which Assistant's answer is the best?} \\[2pt]
\textbf{Final Best Assistant:} “Assistant 1” \ \ or \ \ “Assistant 2” \ \ or \ \ “Equal”
\end{tcolorbox}
\caption{Evaluation prompt used in MILE-RefHumEval for pairwise comparison of two candidate answers (PandaLM Benchmark).}
\label{fig:evaluation_criteria_pairwise_comparison}
\end{figure}

\begin{figure}[h]
\centering
\begin{tcolorbox}[
    colframe=black,
    colback=purple!10,
    boxrule=0.4pt,
    arc=3pt,                
    width=\columnwidth,     
    fontupper=\scriptsize,  
    enhanced
]
\textbf{You are Evaluator "A"}, assigned to assess the quality of 
\textbf{"Caption Generation"} from the \textbf{"Image"}. \\[4pt]
\textbf{[The Start of Caption]} \\
\texttt{\{caption\}} \\
\textbf{[The End of Caption]} \\
\\[-2pt]
\textbf{Is this a good caption for this image?} \\[2pt]
\textbf{Final Score:} (1=good, 0=bad)
\end{tcolorbox}
\caption{Evaluation prompt used in MILE-RefHumEval for the Image Captioning benchmark to assess LLM-generated captions (OID Rated Image Caption Benchmark).}
\label{fig:evaluation_criteria_image_captioning}
\end{figure}


In image captioning (Figure~\ref{fig:evaluation_criteria_image_captioning}), custom prompts assess caption quality relative to images. Similarly, for summarization (Figure~\ref{fig:evaluation_criteria_summarization}), evaluation focuses on coherence, consistency, fluency, and relevance.  In the topical chat task (Figure~\ref{fig:evaluation_criteria_topical_chat}), the evaluation prompt can assess dimensions like contextual understanding, naturalness, coherence, engagement, and groundedness.

The key advantage of MILE-RefHumEval lies in its flexibility and adaptability. Evaluation prompts can be easily customized to suit different tasks by simply redefining the evaluation criteria, without requiring extensive domain expertise. Additionally, evaluators can be dynamically added, removed, or replaced depending on the task requirements. In image captioning, an image-to-text LLM like InternVL can boost evaluation quality.

This framework enables researchers to define any number of evaluation dimensions with detailed criteria tailored to their objectives. Finally, regardless of the task, the final evaluation scores are computed through majority voting across evaluators, which enhances robustness and mitigates individual model bias. However, for tasks where evaluators provide numerical ratings such as summarization (Figure~\ref{fig:evaluation_criteria_summarization}) or other continuous scoring benchmarks, the final score is obtained by averaging the ratings from all evaluators instead of majority voting. By leveraging this decentralized, interaction-free approach, MILE-RefHumEval supports structured, interpretable, and scalable evaluation, even in scenarios where ground-truth reference answers are unavailable.

\begin{figure}[t]
\centering
\begin{tcolorbox}[
    colframe=black,
    colback=green!10,
    boxrule=0.4pt,
    arc=3pt,
    width=\columnwidth,
    fontupper=\scriptsize,
    enhanced
]
\textbf{You are Evaluator "A"}, assigned to assess the quality of 
\textbf{"Summarization"} from the \textbf{"Document text (News Article)"}. \\[4pt]
--- Document Text \& Summary ---
\\ \texttt{\{document\}} \\[4pt]
\textbf{[The Start of Summary]} \\
\texttt{\{summary\}} \\
\textbf{[The End of Summary]} \\
\medskip

\textbf{1. Coherence (1--5)} \\
\textit{Definition:} Logical flow and structure of the summary. A coherent summary is well-organized, with sentences connected naturally and presenting a clear progression of ideas. \\
\textit{Steps:}
1. Read the news article carefully and identify the main topic and key points.
2. Read the summary and check whether it presents information in a logical order without contradictions or abrupt jumps.
3. Assign a score based on overall logical flow.
\textbf{Final Score:} (number between 1 and 5) \\

\textbf{2. Consistency (1--5)} \\
\textit{Definition:} Faithfulness of the summary to the source text. A consistent summary should not contradict, misrepresent, or invent details. \\
\textit{Steps:}
1. Compare the summary’s statements with the source text.
2. Check if the summary accurately reflects the article’s facts without adding false information.
3. Assign a score based on factual accuracy.
\textbf{Final Score:} (number between 1 and 5) \\

\textbf{3. Fluency (1--5)} \\
\textit{Definition:} Readability and grammatical quality of the summary. A fluent summary uses correct grammar, natural phrasing, and smooth sentence transitions. \\
\textit{Steps:}
1. Read the summary carefully, focusing on grammar, spelling, punctuation, and sentence flow.
2. Judge whether the writing is natural and easy to read.
3. Assign a score based on overall fluency.
\textbf{Final Score:} (number between 1 and 5) \\

\textbf{4. Relevance (1--5)} \\
\textit{Definition:} The degree to which the summary focuses on the important and central aspects of the article while avoiding unnecessary or trivial details. \\
\textit{Steps:}
1. Identify the most important points in the article.
2. Check if the summary captures these main points while avoiding irrelevant or minor details.
3. Assign a score based on overall relevance.
\textbf{Final Score:} (number between 1 and 5)
\end{tcolorbox}
\caption{Evaluation prompt used in MILE-RefHumEval for the Summarization benchmark to assess LLM-generated summaries (SummEval Benchmark).}
\label{fig:evaluation_criteria_summarization}
\end{figure}

\begin{figure*}[t]
\centering
\begin{tcolorbox}[
    colframe=black,
    colback=orange!10,
    boxrule=0.4pt,
    arc=7pt,
    width=\textwidth,
    fontupper=\scriptsize,
    enhanced
]
\textbf{You are Evaluator "A"}, assigned to assess the quality of 
\textbf{"Conversational Responses"} from the \textbf{"Topical Chat"} benchmark. \\
--- Conversation \& Candidate Response ---
\\
\textbf{[The Start of Conversation]} \\
\texttt{\{str(context)\}} \\
\textbf{[The End of Conversation]} \\
\textbf{[The Start of Assistant’s Response]} \\
\texttt{\{response\}} \\
\textbf{[The End of Assistant’s Response]} \\
\medskip

\textbf{1. Understandable (0--1)} \\
Is the response understandable in the context of the history? (Not if its on topic, but for example if it uses pronouns they should make sense)\\
-- A score of 0 (no) means that the response is difficult to understand. You do not know what the person is trying to say.
-- A score of 1 (yes) means that the response is understandable. You know what the person is trying to say.
\textbf{Final Score:} (0 = not understandable, 1 = understandable) \\

\textbf{2. Natural (1--3)} \\
Is the response naturally written? \\
\textit{Steps:}
-- A score of 1 (bad) means that the response is unnatural.
-- A score of 2 (ok) means the response is strange, but not entirely unnatural.
-- A score of 3 (good) means that the response is natural.
\textbf{Final Score:} (1 = bad, 2 = okay, 3 = good) \\

\textbf{3. Maintains Context (1--3)} \\
Does the response serve as a valid continuation of the conversation history? \\
-- A score of 1 (no) means that the response drastically changes topic or ignores the conversation history.
-- A score of 2 (somewhat) means the response refers to the conversation history in a limited capacity (e.g., in a generic way) and shifts the conversation topic.
-- A score of 3 (yes) means the response is on topic and strongly acknowledges the conversation history.
\textbf{Final Score:} (1 = off-topic, 2 = partially relevant, 3 = fully on-topic) \\

\textbf{4. Engaging (1--3)} \\
Is the response dull/interesting? \\
-- A score of 1 (dull) means that the response is generic and dull.
-- A score of 2 (somewhat interesting) means the response is somewhat interesting and could engage you in the conversation (e.g., an opinion, thought)
-- A score of 3 (interesting) means the response is very interesting or presents an interesting fact
\textbf{Final Score:} (1 = dull, 2 = somewhat interesting, 3 = very interesting) \\

\textbf{5. Uses Knowledge (0--1)} \\
Given the interesting fact that the response is conditioned on, how well does the response use the fact? \\
-- A score of 0 (no) means the response does not mention or refer to the fact at all
-- A score of 1 (yes) means the response uses the fact well
\textbf{Final Score:} (0 = no, 1 = yes) \\

\textbf{6. Overall Quality (1--5)} \\
Given your answers above, what is your overall impression of this utterance? \\
-- A score of 1 (very bad). A completely invalid response. It would be difficult to recover the conversation after this.
-- A score of 2 (bad). Valid response, but otherwise poor in quality.
-- A score of 3 (neutral) means this response is neither good nor bad. This response has no negative qualities, but no positive ones either.
-- A score of 4 (good) means this is a good response, but falls short of being perfect because of a key flaw.
-- A score of 5 (very good) means this response is good and does not have any strong flaws.
\textbf{Final Score:} (1 = very bad, 2 = bad, 3 = neutral, 4 = good, 5 = very good)

\end{tcolorbox}
\caption{Evaluation prompt used in MILE-RefHumEval for the Topical Chat benchmark to assess LLM-generated conversational responses (Topical Chat Benchmark).}
\label{fig:evaluation_criteria_topical_chat}
\end{figure*}

\section{Experiment}
\subsection{Experimental Setup}

\textbf{FairEval Benchmark}~\citep{Wang2023LargeLM}: The experiments were conducted on 80 samples 
from the Open-ended Question Answers dataset
, comprising nine categories to ensure semantic and domain diversity.
We benchmark MILE-RefHumEval against two baselines: CHATEVAL~\citep{chan2023CHATEVAL}, which employs a single evaluator in a multi-role configuration; and MILE-RefHumEval-Conv, a variant of  conversational ensemble variant where evaluators sequentially review outputs which involves MILE-RefHumEval with inter-evaluator communication as in CHATEVAL.
MILE-RefHumEval uses seven diverse LLMs: DeepSeek-R1-Distill-Llama-8B, Mistral-Small-3.2-24B-Instruct-2506, GPT-4.1-mini, Meta-Llama-3.1-8B-Instruct, Gemma-3-12b-it, Phi-4 and Qwen\_qwq-32b as independent evaluators. These models range from 8B to 32B parameters and were selected to provide complementary architectural and linguistic perspectives. Though MILE-RefHumEval uses two more agents than CHATEVAL, it requires fewer queries.

Evaluation metrics include Accuracy (Acc.), F1-score (F1), Cohen’s Kappa (Kap.), Matthews Correlation Coefficient (MCC), and Query Efficiency to assess reliability, alignment, and cost-effectiveness.

\textbf{SummEval Benchmark} ~\citep{fabbri2021summeval}:
We benchmark MILE-RefHumEval against G-Eval-4~\citep{liu-etal-2023-g} on a summarization task with 1,600 samples 
evaluated on coherence, consistency, fluency, and relevance (scored 1–5). Unlike previous benchmarks, which chose 
the best candidate, MILE-RefHumEval extends this by evaluating fine-grained quality scores that closely approximate human evaluations. For this dataset experiment, we used all seven evaluators, consistent with the setup in the FairEval benchmark.

To assess how accurately MILE-RefHumEval predicts human-like scores, we computed Mean Squared Error (MSE), Root Mean Squared Error (RMSE), and Mean Absolute Error (MAE). These metrics quantify the variance and absolute differences between our predicted scores and human annotations, providing a more precise evaluation beyond simple correlation measures. Since G-Eval-4 
already provides evaluation scores result, we directly use its outputs for benchmarking without rerunning the G-Eval original evaluation process.

\textbf{OID Rated Image Caption Benchmark}~\citep{levinboim2019quality}:
We benchmarked MILE-RefHumEval, a flexible framework for image-to-text evaluation, on the OID Image Caption Quality Estimation Test dataset, which contains 6,182 samples with human-rated captions. Each sample includes an image, a system-generated caption, and the number of positive ratings, from which we computed the rounded average of good ratings.

MILE-RefHumEval adapts easily to diverse tasks by swapping in specialized evaluators. Here, we use five: Mistral-Small-3.2-24B-Instruct-2506, Gemma-3-12B-IT, InternVL3-14B-Instruct, Qwen2.5-VL-7B-Instruct, and GPT-4.1-mini.

To evaluate MILE-RefHumEval’s accuracy, we framed its judgments as a binary classification task: a caption is labeled as “good” if its rounded average positive rating exceeds 0.50 (i.e., more than 2 out of 5 evaluators rated it positively). For comparison, we employed ClipScore~\citep{hessel2021clipscore} (CLIP-ViT-L-14), using a cosine similarity threshold of 0.30 to separate satisfactory captions from unsatisfactory ones. This threshold aligns with recent studies ~\citep{schuhmann2021laion, birhane2023into}, where values around 0.3–0.35 are commonly adopted for zero-shot caption quality evaluation, balancing sensitivity and precision.
Evaluation metrics include Acc., F1, Kap., MCC.

\textbf{PandaLM Benchmark}~\citep{zhu2023judgelm}:
We also compared MILE-RefHumEval to JudgeLM and a PandaLM method specifically trained on the evaluation task dataset. 
The dataset consists of 1,000 samples, each covering diverse categories (e.g., Grammarly, Wikipedia, Coursera,...). Each sample includes an instruction, an input (on various topics), and two responses generated by the same model based on that instruction and input. Three annotators choose the best response.

MILE-RefHumEval use all seven evaluators to select the best response for each sample and then compute the agreement between MILE-RefHumEval’s choice and the human annotations. 
Evaluation metrics include Acc., F1, Kap., MCC.

\textbf{Topical Chat Benchmark}~\citep{mehri-eskenazi-2020-usr}:
To compare MILE-RefHumEval with G-Eval, we used the USR: Dialog Quality Annotations dataset
. This dataset provides annotations across five dimensions: Understandable, Uses Knowledge (0 or 1), Natural, Maintains Context, Engaging (1, 2, or 3), and Overall Quality (1, 2, 3, 4, or 5). We evaluate on the Amazon Topical-Chat datasets (360 samples), with 3 human evaluators.

Evaluation metrics include 
Spearman and Kendall correlation coefficients between MILE-RefHumEval’s judgment scores and those of all three human annotators on Naturalness (
average of understandability and naturalness), Coherence (main context), Groundedness (use of knowledge), and AVG (overall quality). This procedure enables a direct comparison with G-Eval-4, as reported in the original paper.

\subsection{Results}

\begin{table*}[t]
\centering
\resizebox{0.9\textwidth}{!}{%
\begin{tabular}{lccccr}
\hline
\textbf{Evaluator} & \textbf{Acc.(\%)} & \textbf{F1-ma.(\%)} & \textbf{MCC(\%)} & \textbf{Kap.} & \textbf{LLM Queries} \\
\hline
\multicolumn{6}{l}{\textit{\textbf{CHATEVAL: One Evaluator, Many Roles (Cross-Role Agent Conversations)}}} \\
Deepseek & 52.50 & 39.16 & 24.96 & 22.78 &  \\
Mistral  & 45.00 & 33.26 & 14.77 & 12.48 &  \\
GPT      & \textbf{63.75} & 45.87 & 37.66 & 35.84 & 880 (80$\times$11) \\
Llama    & 52.50 & 39.05 & 31.06 & 25.51 &  \\
Gemma    & 57.50 & 41.63 & 27.78 & 26.63 &  \\
Phi      & 55.00 & 43.88 & 23.90 & 23.32 &  \\
Qwen      & 62.50 & 47.79 & 34.62 & 33.05 &  \\
\hline
\multicolumn{6}{l}{\textit{\textbf{MILE-RefHumEval: One Role, Many Evaluators (No Conversations)}}} \\
DeepSeek+Mistral+GPT+Llama+Gemma+Phi+Qwen & \textbf{63.75} & \textbf{49.87} & \textbf{39.04} & \textbf{37.42} & 560 (80$\times$7) \\
\hline
\multicolumn{6}{l}{\textit{\textbf{MILE-RefHumEval-Conv: One Role, Many Evaluators (With Conversations)}}} \\
DeepSeek$\rightarrow$Mistral$\rightarrow$GPT$\rightarrow$Llama$\rightarrow$Gemma$\rightarrow$Phi$\rightarrow$Qwen & 58.75 & 42.84 & 32.66 & 30.38 &  \\
Qwen$\rightarrow$Mistral$\rightarrow$GPT$\rightarrow$Llama$\rightarrow$Gemma$\rightarrow$Phi$\rightarrow$DeepSeek                   & 58.75 & 42.49 & 28.82 & 27.97 &  \\
Phi$\rightarrow$Mistral$\rightarrow$GPT$\rightarrow$Llama$\rightarrow$Gemma$\rightarrow$Qwen$\rightarrow$DeepSeek                  & 58.75 & 42.42 & 30.05 & 28.57 & 1760 (80$\times$22) \\
Llama$\rightarrow$Mistral$\rightarrow$GPT$\rightarrow$Phi$\rightarrow$Gemma$\rightarrow$Qwen$\rightarrow$DeepSeek                   & 53.75 & 38.45 & 19.79 & 18.86 &  \\
GPT$\rightarrow$Mistral$\rightarrow$Qwen$\rightarrow$Llama$\rightarrow$Gemma$\rightarrow$Phi$\rightarrow$DeepSeek                   & 61.25 & 43.67 & 32.51 & 30.80 &  \\
Gemma$\rightarrow$DeepSeek$\rightarrow$GPT$\rightarrow$Llama$\rightarrow$Qwen$\rightarrow$Phi$\rightarrow$Mistral                   & 57.50 & 41.00 & 25.01 & 23.66 &  \\
\hline
\end{tabular}
}
\caption{\label{tab:comparison}
\textbf{Evaluation on FairEval benchmark.} This table compares \textit{CHATEVAL}, a single evaluator performing multiple roles through cross-role agent conversations, and \textit{MILE-RefHumEval-Conv}, representing multi-evaluator conversations, with our \textit{MILE-RefHumEval} approach, employing multiple independent evaluators without interaction.  Metrics \textbf{Accuracy (Acc.\%)}, \textbf{Macro F1-score (F1-ma.\%)}, \textbf{Matthews Correlation Coefficient (MCC\%)}, \textbf{Cohen’s Kappa (Kap.)}, and \textbf{LLM Queries}, illustrate differences in evaluation effectiveness and computational cost.}
\end{table*}
\textbf{FairEval Benchmark Result:}
Table~\ref{tab:comparison} presents the comparison of the three evaluation strategies.
The proposed MILE-RefHumEval approach achieves the highest overall performance. The full ensemble (DeepSeek, Mistral, GPT, LLaMA, Gemma, Phi, and Qwen) reaches 63.75\% accuracy, 49.87\% macro F1, 39.04\% MCC, and 37.42\% Cohen's kappa. This represents an absolute gain of +5\% F1, +1.38\% MCC, and +1.58\% Kap. over CHATEVAL’s best result, while reducing the number of queries by 320. These gains demonstrate the effectiveness of model diversity and independence in reducing bias and improving evaluation reliability.


While CHATEVAL is lightweight in design, it consistently underperforms 
MILE-RefHumEval.
Its best configuration, using GPT as the sole evaluator, yields 63.75\% acc., 45.87\% macro F1, 37.66\% MCC, and a Kap. of 35.84\%. These results indicate moderate alignment with human labels but limited robustness, likely due to role overloading and intra-agent bias. For CHATEVAL, selecting the best evaluator requires running all evaluators, which increases the number of queries. In contrast, for MILE-RefHumEval, we can simply combine the evaluators without increasing the query count.

Under MILE-RefHumEval-Conv, conversational evaluation exhibits significant variability depending on the order in which evaluators are applied, leading to differing performance across configurations. The top-performing variant reaches 61.25\% acc., 43.67\% macro F1, 32.51\% MCC, and 30.80\% kap., with GPT in the leading position. While CHATEVAL achieves its best results with GPT as the evaluator, in MILE-RefHumEval-Conv the highest performance occurs when the conversation is initiated by GPT, indicating that GPT strongly influences the overall evaluation outcome.
This  suggests that cross-evaluator dialogues introduce additional noise and potential bias, leading to inconsistent judgments rather than resolution or consensus.
In terms of computational cost, MILE-RefHumEval is also more efficient. Even with two more agents than CHATEVAL, it requires only 560 LLM queries (80 examples × 7 evaluators), compared to 880 for CHATEVAL (80 × 11 roles) and up to 1760 for MILE-RefHumEval-Conv (80 × 22 steps). Thus, our method not only improves reliability and agreement but also reduces evaluation overhead by 36.4\% relative to CHATEVAL and 68.2\% relative to conversational ensembles.
\begin{figure}[ht]
\centering
  \includegraphics[width=1.1\columnwidth]{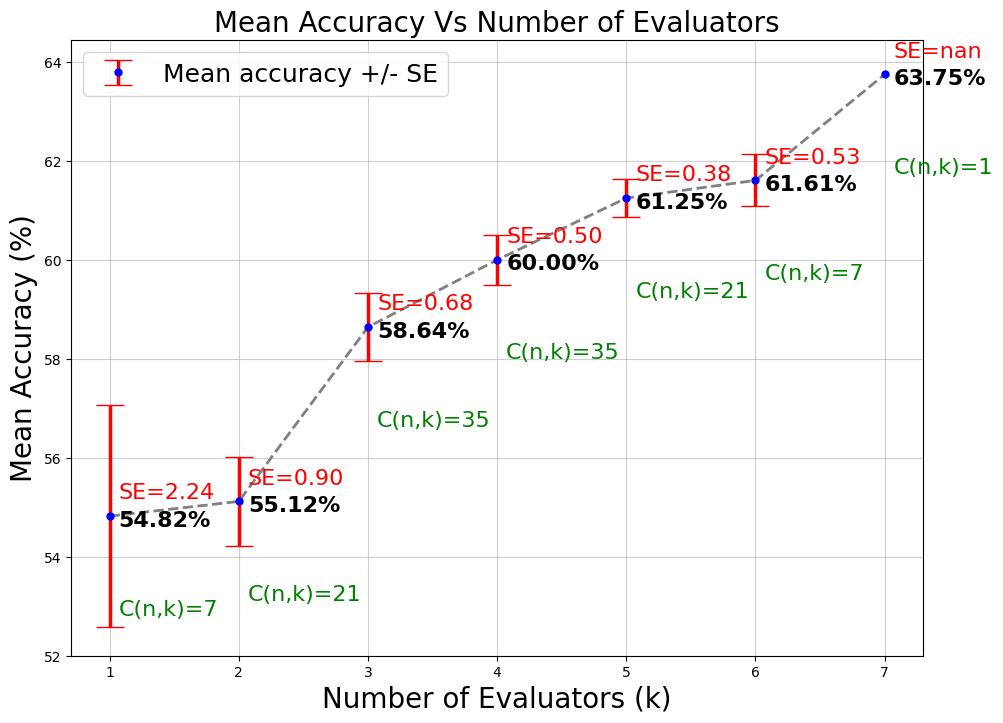}
  \caption{Average accuracy for all possible combinations of $k$ evaluators (binomial coefficient $\binom{N}{k}$).}
  \label{fig:nb_eval_vs_acc}
\end{figure}

Figure~\ref{fig:nb_eval_vs_acc} illustrates how mean accuracy and its associated standard error (SE) evolve as the number of evaluators ($k$) increases. Mean accuracy rises steadily with each additional evaluator, showing that aggregating multiple judgments systematically enhances evaluation reliability. A single evaluator performs poorly (54.82\%), reflecting both low accuracy and high variability. With two evaluators, performance improves only marginally (55.12\%), but beyond this point, the benefits become clear: by $k$=3, accuracy reaches 58.64\%, and at $k$=4, it climbs to 60.00\%. Additional evaluators continue to improve performance, although with diminishing returns, reaching 63.75\% at $k$=7. This trend suggests convergence toward a stable, high-quality consensus with more evaluators.

The behavior of the SE values further clarifies this pattern. SE drops sharply as $k$ increases from 1 to 4 declining from 2.24 at $k$=1 to 0.90 at $k$=2, and stabilizing around 0.50 from $k$=3 onward. This indicates that variability among evaluator subsets decreases rapidly when multiple evaluators are combined. For higher $k$ values, SE remains consistently low (0.38–0.53), reflecting stable estimates, while at $k$=7 it becomes undefined since only one subset exists. When $\binom{n}{k}$ is large such as 35 subsets for $k$=3 and $k$=4 the uncertainty is averaged over many combinations, yielding robust and minimal SE values. Conversely, small subset counts (e.g., $k$=1) lead to more volatile SE due to limited sampling.

Overall, these results demonstrate that increasing the number of evaluators not only improves expected accuracy but also makes poor outcomes less likely to occur by chance. This is especially important in settings where the individual quality or reliability of evaluators is uncertain.

\begin{table}[t]
\centering
\small
\begin{tabular}{lccc}
\hline
\multicolumn{4}{c}{\textbf{MILE-RefHumEval}} \\
\hline
\textbf{Dimension} & \textbf{MSE} & \textbf{RMSE} & \textbf{MAE}  \\
\hline
Coherence & 0.927 & 0.963 & 0.761  \\
Consistency & 0.596 & 0.772 & 0.567  \\
Fluency & 0.641 & 0.801 & 0.685  \\
Relevance & 0.495 & 0.704 & 0.557  \\
\hline
\multicolumn{4}{c}{\textbf{G-Eval-4}} \\
\hline
\textbf{Dimension} & \textbf{MSE} & \textbf{RMSE} & \textbf{MAE}  \\
\hline
Coherence & 1.258 & 1.122 & 0.910  \\
Consistency & 1.849 & 1.360 & 1.083  \\
Fluency & 4.473 & 2.115 & 2.037  \\
Relevance & 0.978 & 0.989 & 0.811  \\
\hline
\end{tabular}
\caption{Performance comparison of MILE-RefHumEval and G-Eval on the SummEval benchmark. Metrics include MSE, RMSE, and MAE across Coherence, Consistency, Fluency, and Relevance. Lower values indicate better agreement with human evaluations.}
\label{tab:summeval_benchmark}
\end{table}

\textbf{SummEval Benchmark Result:}
Table~\ref{tab:summeval_benchmark} shows that MILE-RefHumEval consistently outperforms G-Eval-4 across all evaluation dimensions, achieving substantially lower MSE, RMSE, and MAE scores. For example, MILE-RefHumEval reduces MSE by 26.3\% in Coherence (0.927 vs. 1.258), 67.7\% in Consistency (0.596 vs. 1.849), and an impressive 85.7\% in Fluency (0.641 vs. 4.473). Relevance also improves significantly, with an MSE of 0.495, nearly 50\% lower than G-Eval-4’s 0.978. Moreover, the variance of the proposed approach's 
predictions remains consistently below 1, demonstrating more stable, accurate, and human-aligned evaluations. 

\begin{table}[h!]
\centering
\small
\setlength{\tabcolsep}{5pt} 
\begin{tabular}{lcccc}
\hline
\textbf{Method} & \textbf{Acc.} & \textbf{F1} & \textbf{Kap.} & \textbf{MCC} \\
\hline
MILE-RefHumEval & 58.91 & \textbf{57.97} & \textbf{22.44} & \textbf{26.67} \\
\hline
Mistral   & \textbf{62.42} & 55.35 & 17.41 & 21.28 \\
Gemma     & 56.63 & 55.31 & 18.60 & 22.82 \\
InternVL  & 57.31 & 56.85 & 13.80 & 13.84 \\
QwenVL    & 50.92 & 46.09 & 10.80 & 17.69 \\
GPT       & 56.96 & 54.99 & 19.97 & 26.05 \\
\hline
ClipScore & 43.27 & 30.37 & 0.14 & 2.23 \\
\hline
\end{tabular}
\caption{Evaluation on the OID Rated Image Caption Benchmark. We compare the ensemble \textbf{MILE-RefHumEval} with individual evaluators and the non-LLM baseline \textbf{ClipScore}.}
\label{tab:OID_classification_metrics}
\end{table}

\textbf{OID Rated Image Caption Benchmark Result:}
Table~\ref{tab:OID_classification_metrics} shows that MILE-RefHumEval, an ensemble of Mistral, Gemma, InternVL, QwenVL, and GPT, achieves 58.91\% Accuracy, slightly below Mistral alone (62.42\%), but surpasses all individual evaluators in Macro F1 (57.97\%), kap. (22.44\%), and MCC (26.67\%), indicating more balanced and human-aligned predictions. Single evaluator like Mistral excel in raw Accuracy, while InternVL maintains strong class-wise performance (Macro F1 56.85\%). QwenVL underperforms across metrics, and ClipScore serves as a low baseline (Accuracy 43.27\%, Kappa 0.14\%). Notably, even on image captioning tasks, the ensemble achieves superior overall performance under the same evaluation criteria, demonstrating that combining evaluators leads to more reliable and robust results than relying on any single evaluator alone. These results highlight that ensembling evaluators improves reliability and alignment with human judgments.
\begin{table}[h!]
\centering
\renewcommand{\arraystretch}{0.9}
\setlength{\tabcolsep}{3pt}
\small
\begin{tabular}{lcccc}
\hline
\textbf{Method} & \textbf{Acc. (\%)} & \textbf{F1 (\%)} & \textbf{Kap.} & \textbf{MCC (\%)} \\
\hline
\multicolumn{5}{c}{\textbf{\textit{Annotator 1}}} \\
MILE-RefHumEval  & \textbf{77.78} & \textbf{70.76} & \textbf{61.43} & \textbf{61.52} \\
PandaLM-7B  & 65.97 & 56.03 & 41.91 & 41.92 \\
JudgeLM-7B & 57.46 & 39.79 & 23.69 & 26.40 \\
\hline
\multicolumn{5}{c}{\textbf{\textit{Annotator 2}}} \\
MILE-RefHumEval  & \textbf{76.98} & \textbf{69.76} & \textbf{60.54} & \textbf{60.73} \\
PandaLM-7B & 65.17 & 56.24 & 41.19 & 41.20 \\
JudgeLM-7B & 56.66 & 39.69 & 23.59 & 26.45 \\
\hline
\multicolumn{5}{c}{\textbf{\textit{Annotator 3}}} \\
MILE-RefHumEval  & \textbf{78.48} & \textbf{71.58} & \textbf{63.04} & \textbf{63.26} \\
PandaLM-7B  & 66.97 & 58.55 & 44.12 & 44.15 \\
JudgeLM-7B & 56.16 & 39.31 & 22.88 & 25.78 \\
\hline
\multicolumn{5}{c}{\textbf{\textit{Human Annotator Majority Vote}}} \\
MILE-RefHumEval & \textbf{78.98} & \textbf{71.90} & \textbf{63.71} & \textbf{63.85} \\
PandaLM-7B  & 66.77 & 57.43 & 43.54 & 43.55 \\
JudgeLM-7B & 57.76 & 40.24 & 24.82 & 27.75 \\
\hline
\end{tabular}
\caption{Evaluation on the PandaLM benchmark using three human annotators and their majority vote. We compare \textbf{MILE-RefHumEval}, \textbf{PandaLM-7B}, and \textbf{JudgeLM-7B} across Accuracy, Macro F1, kap., and MCC. PandaLM-7B results are computed from the official repository outputs, while JudgeLM-7B was downloaded and evaluated locally using the same settings as MILE-RefHumEval.}
\label{tab:pandalm_results}
\end{table}

\textbf{PandaLM Benchmark Result:}
Table~\ref{tab:pandalm_results} reports the performance of MILE-RefHumEval, PandaLM-7B, and JudgeLM-7B across three independent annotators and their majority vote on the PandaLM benchmark. MILE-RefHumEval consistently outperforms both baselines across all metrics and annotators, achieving the highest overall performance with an average Acc. of 78.98\%, Macro F1 of 71.90\%, Kap. of 63.71\%, and MCC of 63.85\% under majority voting. In comparison, PandaLM-7B performs moderately well (Accuracy 66.77\%, F1 57.43\%), while JudgeLM-7B lags significantly behind (Accuracy 57.76\%, F1 40.24\%).

Across individual annotators, MILE-RefHumEval maintains consistently high Accuracy (76.98–78.48\%) and Macro F1 (69.76–71.58\%), showing strong robustness to annotator disagreement. The improvements over PandaLM-7B are substantial, averaging +12.2 points in Accuracy, +14.5 in Macro F1, and +20.2 in kap., demonstrating significantly stronger agreement with human judgments. JudgeLM-7B’s low Kap. scores (22.88–26.45\%) confirm its weaker alignment with human preferences.
\begin{table*}[t]
\centering
\small
\setlength{\tabcolsep}{4pt}
\renewcommand{\arraystretch}{1.0}
\begin{tabular}{lcccccccccc}
\hline
\textbf{Method} & \multicolumn{2}{c}{\textbf{Naturalness}} & \multicolumn{2}{c}{\textbf{Coherence}} & \multicolumn{2}{c}{\textbf{Engagingness}} & \multicolumn{2}{c}{\textbf{Groundedness}} &  \multicolumn{2}{c}{\textbf{AVG}} \\
\cline{2-11}
 & r & $\rho$ & r & $\rho$ & r & $\rho$ & r & $\rho$ & r & $\rho$ \\
\hline
MILE-RefHumEval & \textbf{0.806} & \textbf{0.750} & \textbf{0.805} & \textbf{0.739} & \textbf{0.746} & \textbf{0.683} & \textbf{0.863} & \textbf{0.813} & \textbf{0.692} & \textbf{0.612} \\
G-Eval* & 0.549 &  0.565 & 0.594 &  0.605 & 0.627 &  0.631 & 0.531 & 0.551 & 0.575 & 0.588 \\
\hline
\end{tabular}
\caption{Average Correlation Scores for MILE-RefHumEval Across Dimensions on the Amazon Topical-Chat benchmark. *G-Eval correlation score is directly from ~\cite{liu-etal-2023-g}}
\label{tab:mile_geval_human_correlations}
\end{table*}

\textbf{Amazon Topical-Chat Benchmark Result:}
Table~\ref{tab:mile_geval_human_correlations} shows MILE-RefHumEval outperforms G-Eval across all dialogue evaluation dimensions, yielding the highest correlations with human judgments.
Specifically, compared to G-Eval, MILE-RefHumEval improves Naturalness correlations by r = 0.257 and $\rho$ = 0.185, Coherence by r = 0.211 and $\rho$ = 0.134, Engagingness by r = 0.119 and $\rho$ = 0.052, and Groundedness by r = 0.332 and $\rho$ = 0.262, respectively, demonstrating clear gains in capturing naturalness, coherence, engagement, and factual consistency.

On average, MILE-RefHumEval achieves r = 0.117 and $\rho$ = 0.024 improvements over G-Eval, confirming that our framework provides the most reliable and human-aligned evaluations across all dimensions. These results strongly validate MILE-RefHumEval as the state-of-the-art method for reference-free dialogue evaluation.

\section{Discussion}
Our results reveal a surprising yet consistent trend: independent, non-conversational evaluator ensembles outperform conversational configurations across all metrics (Table~\ref{tab:comparison}). While conversational paradigms are often assumed to emulate human collaborative reasoning, our analysis shows they introduce inter-model bias, where earlier model outputs unduly influence subsequent evaluators. This undermines objectivity and reduces alignment with human judgments. Crucially, the order and how much it strong influence in which models participate in conversations significantly affects outcomes.
\\
\textbf{Diverisity matters more than size.} The results suggest that diversity of evaluators contributes more to human alignment than simply using a larger or stronger single LLM. For example, while GPT alone achieved high accuracy in some tasks, the ensemble’s balanced performance across dimensions like coherence, fluency, and engagement shows that a mix of reasoning styles outperforms raw scale. This implies that evaluation reliability stems not from model strength alone, but from the complementarity of evaluators.
\\
\textbf{Order sensitivity exposes hidden fragility in conversational evaluators.} The poor stability of conversational ensembles reveals an overlooked issue: the order in which evaluators “speak” acts as a hidden variable that can sway results dramatically. This is analogous to psychological priming in human group discussions, where the first speaker heavily shapes consensus. By removing order effects, MILE-RefHumEval uncovers that evaluator independence is essential for trustworthy automation a point rarely considered in prior benchmarks.
\\
\textbf{Fluency is the hardest dimension for evaluators to approximate.} In the SummEval results, fluency showed the largest error reduction (85.7\%) compared to baselines, suggesting that models struggle most with accurately judging naturalness and readability. That MILE-RefHumEval narrowed this gap highlights that diverse evaluators collectively capture surface-level language features better than individual judges. 
\\
\textbf{Single models can “spike” in performance, but ensembles stabilize.} In the Image Caption benchmark, Mistral alone slightly outperformed the ensemble in raw accuracy, but its gains came at the cost of much lower agreement metrics. This finding mirrors the idea of “overfitting” in supervised learning where one evaluator might excel on certain samples but lack generalizable reliability. Ensembles act as a variance-reduction mechanism, producing more human-like stability even if they sometimes underperform on headline accuracy.
\\
\textbf{Where humans disagree, ensemble evaluators thrive.} In PandaLM, human annotators disagreed on many samples, yet MILE-RefHumEval maintained consistently high agreement across annotators. This suggests that ensembles may not just replicate human judgment, but actually mediate between conflicting human opinions, offering a form of consensus-building absent in individual LLM judges. That makes ensembles not just a human-aligned tool, but potentially a human-reflective evaluator capturing the “average” of human perspectives more reliably than any one person.
\section{Conclusion}
MILE-RefHumEval introduces a reference-free simple yet powerful shift in LLM evaluation: independence over interaction. By leveraging diverse, non-conversational ensembles, the framework reduces bias, stabilizes judgments, and achieves stronger alignment with human evaluations across tasks and modalities. Our findings highlight that evaluation quality stems not from mimicking human debate, but from diversity, independence, and structured criteria. As such, MILE-RefHumEval offers a scalable and reliable foundation for future human-aligned assessment of LLMs.

Looking ahead, the benchmark should be extended to include audio and speech-based tasks (e.g., MMAU benchmark~\cite{sakshi2024mmau} for sound, music, and other audio modalities). Incorporating specialized evaluators such as Audio Flamingo 3, Qwen 2.5 Omni, and Gemini 2.5 Pro will help determine whether these enhancements improve evaluation reliability. Identifying how many evaluators maximize accuracy without redundancy is an important future direction.

\newpage

\subsection*{Limitations}
\paragraph{Evaluator Generalization:} Our framework assumes uniform evaluator reliability across tasks, yet LLMs often exhibit domain-specific capabilities. Without mechanisms for domain-aware weighting or specialization, our current design may miss contextual subtleties in judgment, limiting its generalization across varied evaluation domains.

\paragraph{Interaction Bias:} The fixed turn order in our conversational protocols introduces directional bias, where early-stage outputs disproportionately shape subsequent evaluations. This effect is especially pronounced when weaker models initiate the sequence. While we quantify this cascading influence, we do not explore adaptive, randomized, or model-quality-aware turn ordering as potential mitigation strategies.

\paragraph{Metric Granularity:} We rely on standard agreement metrics (e.g., Pearson correlation, kap.), which may fail to capture nuanced shifts in evaluative reasoning introduced through interaction. Richer analytical tools such as causal attribution methods, latent trajectory comparisons, or fine-grained human annotations are needed to trace how evaluators influence one another.

\paragraph{Prompt Sensitivity:} We observe significant variation in model behavior across prompt formulations, particularly among smaller models like Phi and Mistral. This sensitivity challenges reproducibility and points to the need for prompt-invariant evaluation strategies or dynamic prompt calibration mechanisms to ensure fairness and stability.

\paragraph{Unexplored Optimal Evaluator Count:} We have not studied how many evaluators are needed to achieve maximal accuracy without introducing redundancy. Determining this optimal ensemble size remains an open question for future work.

\paragraph{Domain Generalization:} MILE-RefHumEval has been tested primarily on general-purpose tasks such as summarization, captioning, and dialogue. Its reliability in highly specialized domains (e.g., legal, medical, or technical reasoning) remains untested, and model evaluators trained on general data may misjudge domain-specific quality.

\bibliography{custom}

\clearpage
\appendix
\onecolumn
\section{Appendices}
\label{sec:appendix}

\begin{figure*}[ht!]
\centering
\begin{tcolorbox}[
    colframe=black,
    colback=gray!10,
    boxrule=0.5pt,
    arc=4pt,
    width=\textwidth,
    fontupper=\footnotesize,
    sharp corners=south,
    enhanced
]
Please evaluate the accuracy, clarity, and relevance of the answer generated for the following question. Provide percentage scores along with detailed explanations for the following criteria: \\
\textbf{1. Answer Relevance:} \\
\hspace*{0.5cm} -  \textbf{Score:} [Provide score between 0 and 100] \\
\hspace*{0.5cm} - \textbf{Explanation:} [Assess the clarity of the response. Is the answer easy to understand and logically structured? Are there any ambiguities or confusing phrases that could hinder comprehension?] \\
\textbf{2. Depth and Completeness:} \\
\hspace*{0.5cm} - \textbf{Score:} [Provide score between 0 and 100] \\
\hspace*{0.5cm} - \textbf{Explanation:} [Evaluate the depth of the response. Does it provide a thorough explanation, or does it lack detail? Are all key aspects of the question addressed adequately?] \\
\textbf{3. Grammar and Linguistic Accuracy:} \\
\hspace*{0.5cm} - \textbf{Score:} [Provide score between 0 and 100] \\
\hspace*{0.5cm} - \textbf{Explanation:} [Assess the grammatical correctness of the response. Are there any spelling or punctuation errors? Is the sentence structure correct and appropriate for the context?] \\
\textbf{4. Contextual Appropriateness:} \\
\hspace*{0.5cm} - \textbf{Score:} [Provide score between 0 and 100] \\
\hspace*{0.5cm} - \textbf{Explanation:} [Evaluate how well the answer adapts to the specific question category (e.g., “generic” or “knowledge”). Does it use relevant terminology and concepts appropriate to the subject matter?] \\
\textbf{5. Conciseness and Precision:} \\
\hspace*{0.5cm} - \textbf{Score:} [Provide score between 0 and 100] \\
\hspace*{0.5cm} - \textbf{Explanation:} [Evaluate whether the response is concise and to the point. Does the answer provide the necessary details without unnecessary elaboration or repetition?] \\
\textbf{6. Creativity and Insight:} \\
\hspace*{0.5cm} - \textbf{Score:} [Provide score between 0 and 100] \\
\hspace*{0.5cm} - \textbf{Explanation:} [Assess the creativity of the response. Does it offer a unique perspective or innovative solution, especially when dealing with complex or thought-provoking questions?] \\
\textbf{7. Bias and Fairness:} \\
\hspace*{0.5cm} - \textbf{Score:} [Provide score between 0 and 100] \\
\hspace*{0.5cm} - \textbf{Explanation:} [Evaluate the response for any signs of bias or unfairness. Does the answer exhibit neutrality, or does it reflect a particular perspective that may be considered biased or imbalanced?] \\
\textbf{8. Knowledge Accuracy:} \\
\hspace*{0.5cm} - \textbf{Score:} [Provide score between 0 and 100] \\
\hspace*{0.5cm} - \textbf{Explanation:} [Assess the accuracy of factual information presented in the answer. Are the facts correct? Are any misconceptions, errors, or outdated information included?] \\
\textbf{9. Adaptability to Various Categories:} \\
\hspace*{0.5cm} - \textbf{Score:} [Provide score between 0 and 100] \\
\hspace*{0.5cm} - \textbf{Explanation:} [Evaluate how well the answer adapts to different categories (e.g., “generic” vs. “knowledge”). Does the response adjust its tone, depth, and complexity based on the question category?] \\
\textbf{10. Scalability of Responses:} \\
\hspace*{0.5cm} - \textbf{Score:} [Provide score between 0 and 100] \\
\hspace*{0.5cm} - \textbf{Explanation:} [Evaluate how well the model handles a variety of question types and complexities. Does the model handle simple and complex questions effectively, or does its performance degrade with more challenging queries?] \\
\textbf{11. Overall Answer Quality:} \\
\hspace*{0.5cm} - \textbf{Score:} [Provide score between 0 and 100] \\
\hspace*{0.5cm} - \textbf{Explanation:} [Provide a final summary explanation for the overall score (Explain how it is calculated to show the calculation.) Consider how well all individual criteria performed. Offer insight into the general quality of the response based on the evaluation results.] \\
\textbf{12. Error Detection and Correction:} \\
Please analyze the provided response and identify any errors, inconsistencies, or areas where the response could be improved. The model should help by suggesting a corrected version of the output, focusing on the following points: \\
\hspace*{0.5cm} - Identifying any missing or inaccurate information. \\
\hspace*{0.5cm} - Correcting any grammatical or linguistic errors. \\
\hspace*{0.5cm} - Ensuring the relevance and completeness of the answer. \\
\hspace*{0.5cm} - Suggesting any improvements in clarity or coherence. \\
\hspace*{0.5cm} - Making the answer more concise and precise where necessary. \\
\textbf{13. Corrected Version:} \\
\hspace*{0.5cm} - \textbf{Updated Answer:} [Provide a corrected, revised version of the answer based on the analysis above.]
\end{tcolorbox}
\caption{Evaluation prompt on FairEval Benchmark used in MILE-RefHumEval to assess LLM-generated answers. The prompt guides each evaluator to score responses across 12 dimensions, including relevance, completeness, grammar, factual accuracy, bias, \textbf{**Overall Answer Quality (Automatically computed by each evaluator as the average across the 10 preceding criteria)}, and finally provide justification and revisions. This structured criteria ensures consistent, multi-faceted evaluation without human intervention.}
\label{fig:evaluation_criteria}
\end{figure*}

\begin{figure*}[t]
  \centering
  \hspace*{-2cm} 
  \includegraphics[width=1.2\textwidth]{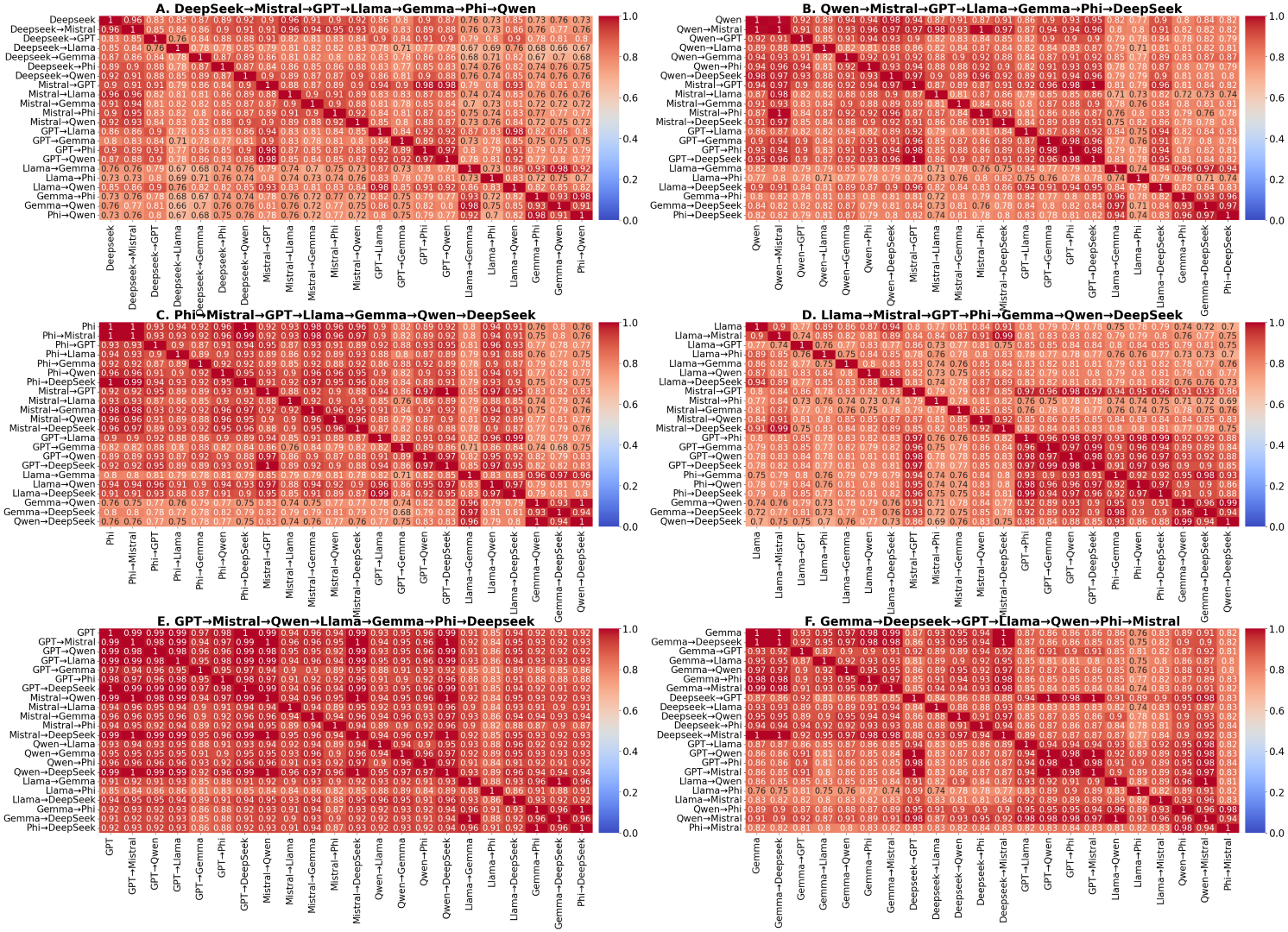}
  \caption{Pearson correlation between multiple LLM evaluators under MILE-RefHumEval-Conv. The figure highlights varying levels of agreement depending on model order and interaction style, suggesting that conversational sequences introduce inconsistency.}
  \label{fig:kappa_heatmap_cross_evaluator_single_role}
\end{figure*}

\begin{figure*}[t]
    \centering
    \hspace*{-2cm} 
    \includegraphics[width=1.2\textwidth]{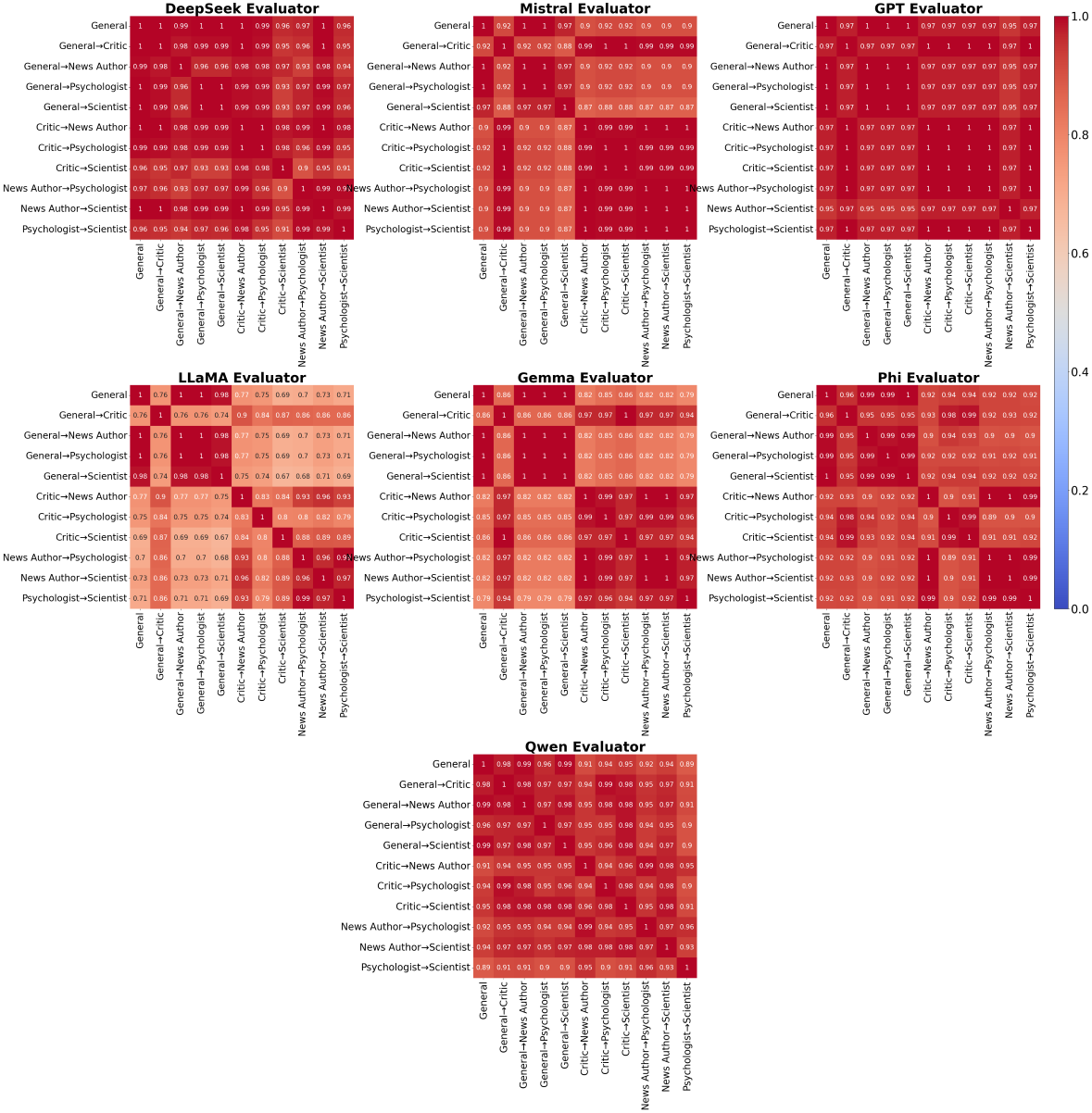}
    \caption{Pearson correlation produced by a single evaluator (CHATEVAL) across different role-played personas. This figure demonstrates that repeating the same model across multiple roles can inflate agreement artificially, reducing diversity of judgment and interpretive value.}
    \label{fig:pearson-agent-roles}
\end{figure*}

\begin{figure*}[t]
    \centering
    \includegraphics[width=0.7\textwidth]{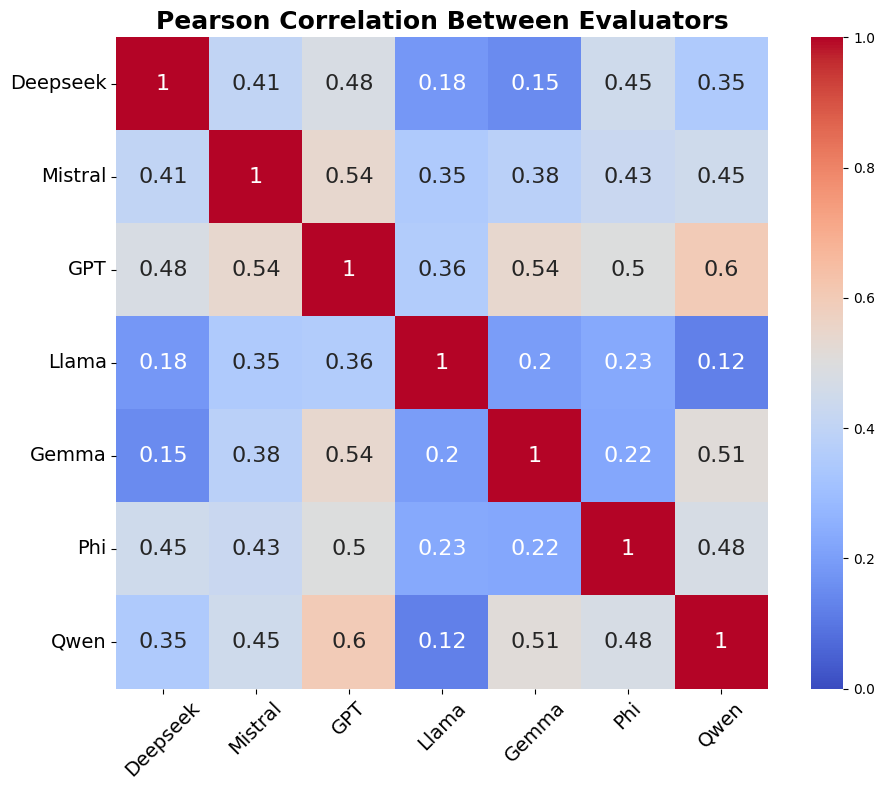}
    \caption{Inter-evaluator agreement analysis in MILE-RefHumEval, showing variance and correlation among independently operating LLMs. The figure emphasizes that non-conversational, diverse evaluators lead to broader reasoning coverage and stronger alignment with human assessments compared to role-based or interactive setups.}
    \label{fig:evaluator-agreement-our-method}
\end{figure*}

\begin{figure*}[t]
    \centering
    \includegraphics[width=\linewidth]{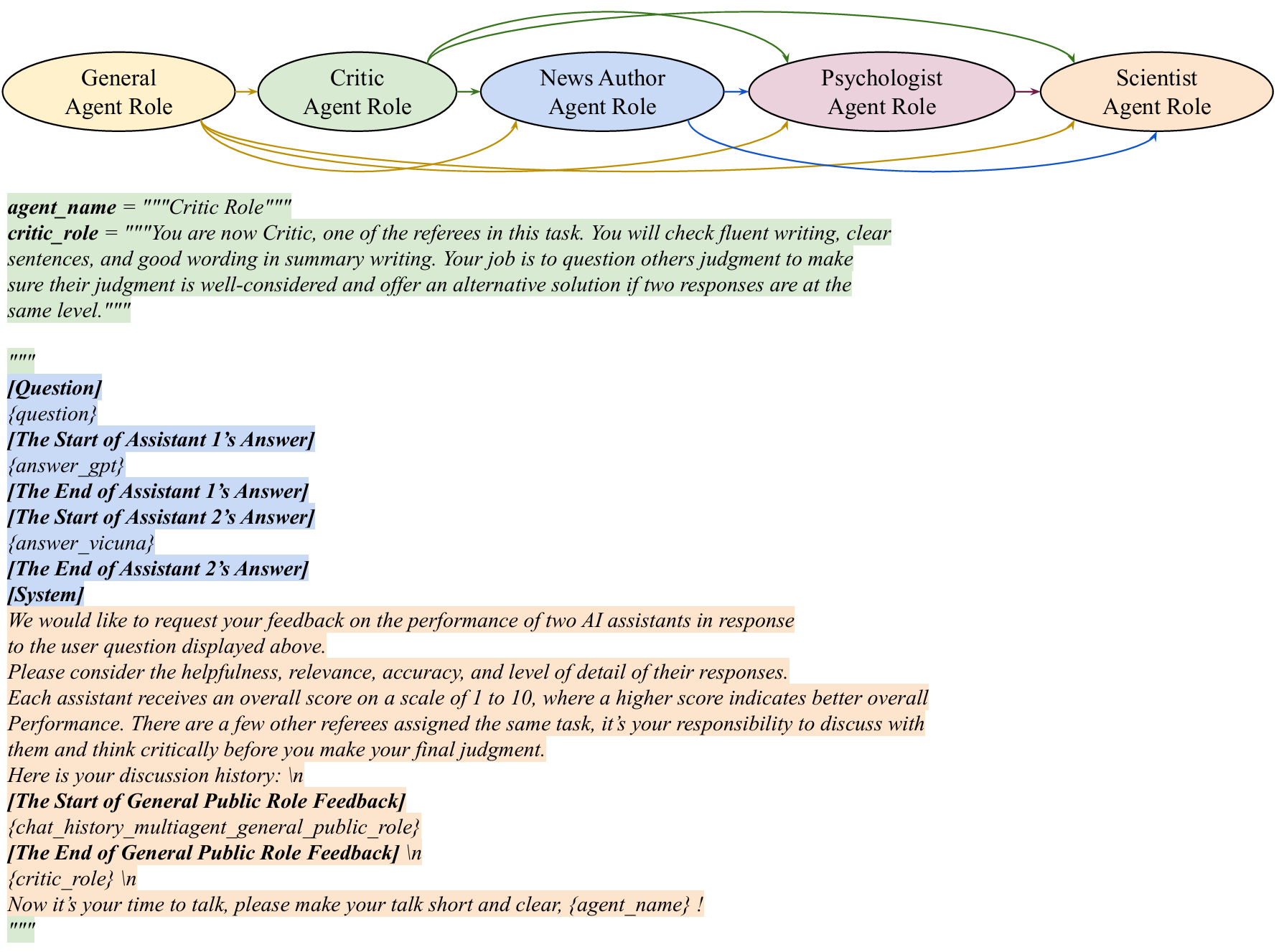}
    \caption{Illustration of CHATEVAL, a baseline evaluation method where a single LLM evaluator sequentially adopts multiple roles (e.g., critic, scientist) to assess responses. This approach introduces potential role contamination and biases due to repeated model use across evaluative stages.}
    \label{fig:baseline_method_CHATEVAL}
\end{figure*}
\begin{figure*}[t]
    \centering
    \includegraphics[width=\linewidth]{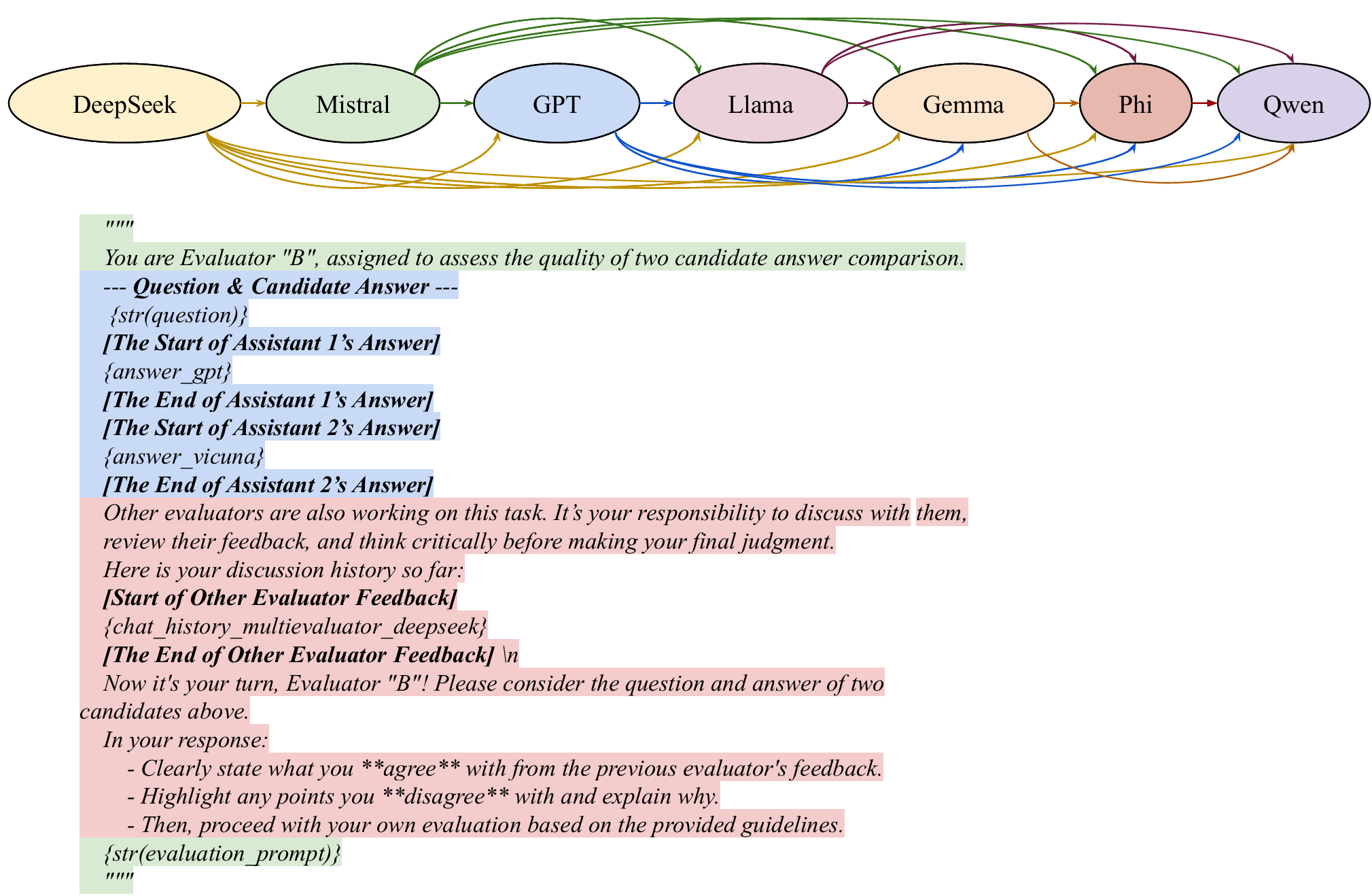}
    \caption{Diagram of MILE-RefHumEval-Conv, where multiple LLM evaluators engage in a conversation chain, each responding to the previous model’s input. This method mimics human dialogue but risks accumulating bias from earlier turns, affecting independent judgment.}
    \label{fig:baseline_method_11}
\end{figure*}

\begin{figure*}[t]
\centering
\begin{tcolorbox}[
    colframe=black,
    colback=orange!10,
    boxrule=0.5pt,
    arc=4pt,
    width=\textwidth,
    fontupper=\footnotesize,
    sharp corners=south,
    enhanced
]
\centering
\makebox[\textwidth]{\textbf{Question:} How can I improve my time management skills?}
\vspace{0.5em} 

\noindent
\begin{minipage}[t]{0.47\textwidth}

\textbf{Assistant 1's answer (GPT):} \\[2pt]
 Here are some tips to improve your time management skills:\\1. Create a schedule: Make a to-do list for the day, week or month and prioritize tasks by importance and deadline.\\2. Set realistic goals: Be realistic about what you can accomplish in a given amount of time. Set goals that can be achieved but are also challenging.\\3. Minimize distractions: Minimize distractions such as emails, phone calls, and social media notifications. Turn off notifications or use apps that block these distractions.\\4. Take breaks: Taking a break can help you recharge and increase productivity. Take a break every hour or so to stretch, go for a walk, or listen to some music.\\5. Learn to say 'no': Learn to say 'no' to tasks and requests that are not a priority or are not within your capabilities.\\6. Delegate tasks: If possible, delegate tasks to others who are capable of completing them. This will help to reduce your workload.\\7. Prioritize self-care: Make sure to take care of yourself by prioritizing self-care activities such as exercise, meditation or spending time with friends and family.\\Remember, time management is a skill that takes time and practice to develop. \\[2pt]
\end{minipage}
\hfill
\begin{minipage}[t]{0.47\textwidth}
\textbf{Assistant 2's answer (VICUNA):} \\[2pt]
 Improving your time management skills can help you to be more productive, focused, and less stressed. Here are some tips to help you improve your time management skills:\\1. Set clear goals: Identify what you want to achieve and set specific, measurable, and realistic goals. Prioritize your tasks and focus on the most important ones first.\\2. Use a calendar or planner: Use a tool to organize your time and schedule your tasks. This can help you to stay on track and make the most of your time.\\3. Break down tasks: Break down larger tasks into smaller, more manageable steps. This can help you to focus on one step at a time and make steady progress.\\4. Eliminate distractions: Identify and eliminate distractions that may be preventing you from being productive. This can include turning off your phone, closing unnecessary tabs on your computer, or finding a quiet place to work.\\5. Practice self-discipline: Use self-discipline to stay focused and avoid procrastination. Set a timer for each task and challenge yourself to complete it before the timer goes off.\\6. Take breaks: Take regular breaks to rest and recharge. This can help you to stay focused and avoid burnout.\\7. Review your progress: Regularly review your progress and adjust your strategy as needed. This can help you to stay on track and make improvements over time. \\[2pt]
\end{minipage}
\end{tcolorbox}
\caption{
Assistant responses to the question "How can I improve my time management skills?" Two LLMs (GPT and VICUNA) provide advice independently. The figure displays these responses side-by-side to support qualitative comparison of tone, clarity, structure, and depth.
}
\label{fig:assistant_comparison_time_management}
\end{figure*}

\begin{figure*}[t]
\centering
\begin{tcolorbox}[
    colframe=black,
    colback=blue!10,
    boxrule=0.5pt,
    arc=4pt,
    width=\textwidth,
    fontupper=\footnotesize,
    sharp corners=south,
    enhanced
]

\makebox[\textwidth]{\textbf{Candidates Evaluation using CHATEVAL: One Evaluator (DeepSeek), Many Roles (Cross-Role Agent Conversations)}}
\vspace{0.5em} 

\hspace*{0.5cm} -  \textbf{General Public Role:} <think> Okay, so I need to figure out which of these two AI assistants did a better job answering the question about improving time management skills. Let me start by reading both responses carefully.Assistant 1's answer has seven tips: creating a schedule with prioritization, setting realistic goals, minimizing distractions, taking breaks, learning to say no, delegating tasks, and prioritizing self-care. It seems pretty comprehensive, covering both personal productivity and some strategic aspects like delegation and goal-setting.On the other hand, Assistant 2 also lists seven tips but they're slightly different. They mention setting clear goals, using a calendar or planner, breaking down tasks into smaller steps, eliminating distractions, practicing self-discipline with timers, taking breaks, and reviewing progress. This approach is more about structure and discipline, focusing on tools like planners and timers.I think both assistants are helpful, but I notice that Assistant 1 includes tips related to self-care and delegating, which might be more beneficial for someone managing multiple responsibilities or needing support. Meanwhile, Assistant 2's emphasis on using tools and setting clear goals could be very practical for someone looking to organize their schedule effectively.\\However, considering the content, both are quite similar in structure but maybe differ slightly in depth. I don't see any major inaccuracies in either response. They both cover essential time management strategies, so it's a bit tricky to choose one over the other.\\But if I have to pick the better one, perhaps Assistant 1 is slightly more comprehensive because they include self-care and delegating, which are important aspects that can help in long-term productivity and reducing stress. Alternatively, maybe Assistant 2's use of specific tools like planners makes their advice more actionable.\\Wait, but both are equally good in terms of structure and relevance. Maybe it's a tie? Or perhaps one is better than the other based on certain criteria. Since I have to choose one, I might lean towards Assistant 1 because they include self-care, which is crucial for maintaining motivation and reducing burnout. </think>\\ \hspace*{1cm}- Final Score for Assistant 1: 9\\ \hspace*{1cm}- Final Score for Assistant 2: 8.5 \\
\hspace*{1cm}- \textcolor{red}{Final Best Assistant: Assistant 1} \\
\hspace*{0.5cm} - \textbf{General Public Role to Critic Role:}\\ \hspace*{1cm}- Final Score for Assistant 1: 9\\ \hspace*{1cm}- Final Score for Assistant 2: 8.5\\ \hspace*{1cm}- \textcolor{red}{Final Best Assistant: Assistant 1} \\
\hspace*{0.5cm} - \textbf{General Public Role to News Author Role:}\\ \hspace*{1cm}- Final Score for Assistant 1: 9\\ \hspace*{1cm}- Final Score for Assistant 2: 8.5\\ \hspace*{1cm}- \textcolor{red}{Final Best Assistant: Assistant 1} \\
\hspace*{0.5cm} - \textbf{General Public Role to Psychologist Role:}\\ \hspace*{1cm}- Final Score for Assistant 1: 9\\ \hspace*{1cm}- Final Score for Assistant 2: 8.5 \\ \hspace*{1cm}- \textcolor{red}{Final Best Assistant: Assistant 1} \\
\hspace*{0.5cm} - \textbf{General Public Role to Scientist Role:}\\ \hspace*{1cm}- Final Score for Assistant 1: 9\\ \hspace*{1cm}- Final Score for Assistant 2: 8.5\\ \hspace*{1cm}- \textcolor{red}{Final Best Assistant: Assistant 1} \\

\end{tcolorbox}
\caption{
Evaluation of assistant responses using CHATEVAL (one LLM acting across multiple roles). Each role (e.g., Critic, Psychologist) scores the same responses using a shared criteria. The figure highlights score consistency across roles but potential redundancy due to a single model’s repeated use.
}
\label{fig:CHATEVAL_role_based_assessment}

\end{figure*}

\begin{figure*}[t]
\centering
\begin{tcolorbox}[
    colframe=black,
    colback=yellow!10,
    boxrule=0.5pt,
    arc=4pt,
    width=\textwidth,
    fontupper=\footnotesize,
    sharp corners=south,
    enhanced
]

\makebox[\textwidth]{\textbf{Candidates Evaluation using MILE-RefHumEval-Conv: (One Role, Many Evaluators (With Conversations))}}
\vspace{0.5em} 

\hspace*{0.5cm} - \textbf{DeepSeek:} <think> Okay, so I need to evaluate the answers from two candidates for the question How can I improve my time management skills? The evaluation criteria are quite detailed with 12 points each, but I'll try to break it down step by step.First, let's look at Assistant 1's answer. It starts by listing tips like creating a schedule and setting realistic goals. Then it goes on to minimize distractions, take breaks, say no, delegate tasks, and prioritize self-care. The answer seems structured with clear points, each starting with a number. It ends with a note that time management takes practice.Now, Assistant 2's answer is similar but phrased differently. They talk about setting clear goals, using a calendar, breaking down tasks, eliminating distractions, practicing self-discipline, taking breaks, reviewing progress, and being adaptable. Their structure is also numbered, and they emphasize the importance of regular reviews.Starting with \textbf{Answer Relevance}: Both answers are directly addressing how to improve time management, so they're very relevant. No off-topic details here. I'll give both a high score, maybe 95 each. \textbf{Clarity and Coherence}: Assistant 1's answer is clear but a bit repetitive in structure. Each point is straightforward, but it might lack some depth. Assistant 2's answer is also clear but more concise. They use similar points but present them differently. I think both are coherent, so another high score, maybe 90 each. \textbf{Depth and Completeness}: Both answers cover common time management tips, which is good. However, they don't go into much detail beyond the basics. Assistant 1 mentions taking breaks and self-care, while Assistant 2 adds using a calendar and breaking down tasks. They both miss some aspects like prioritizing tasks more effectively or dealing with unexpected obstacles. I'll give them an 85 each. \textbf{Grammar and Linguistic Accuracy}: Both answers are grammatically correct. No spelling errors noticed. The sentences are clear and appropriate for the context. Maybe a solid 90 each. \textbf{Contextual Appropriateness}: Since the question is about time management, both answers use relevant terminology without unnecessary jargon. They're appropriate for the category. I'll score this as 85 each. \textbf{Conciseness and Precision}: Assistant 1's answer is concise but could be more precise by elaborating on each point a bit. Assistant 2's answer is also concise but perhaps too brief in some areas. Both are pretty good, so maybe 90 each. \textbf{Creativity and Insight}: Neither answer provides unique perspectives; they stick to common advice. There's no innovative solution here. I'll give them both 75. \textbf{Bias and Fairness}: No signs of bias. Both answers are neutral and balanced. So, 100 each. \textbf{Knowledge Accuracy}: The tips provided are accurate for time management. No errors detected. Both get 100. \textbf{Adaptability to Various Categories}: Both answers are suitable for the question category. They don't need to change much, so 90 each. \textbf{Scalability of Responses}: Both handle simple questions well but might struggle with more complex ones. Maybe a score of 75 each. \textbf{Overall Answer Quality}: Considering all criteria, both are good but not excellent. I'll give them an average score, say 80 each. </think> - Final Score for Assistant 1: 80 - Final Score for Assistant 2: 80 - \textcolor{red}{Final Best Assistant: Equal}  \\

\hspace*{0.5cm} - \textbf{DeepSeek to Mistral:} \textbf{Agree}:- I agree with the other evaluator's assessment that both answers are relevant and neutral.- Both answers score high in grammar and linguistic accuracy. \textbf{Disagree}:- I disagree slightly with the conciseness and precision scores. Assistant 1's answer is a bit repetitive, and Assistant 2's answer could benefit from more elaboration on some points.- I also disagree with the scalability score; both answers handle simple questions well but might struggle with more complex queries. 1. \textbf{**Answer Relevance**}:- **Score**: 95- **Explanation**: Both answers directly address how to improve time management skills and stay on topic without any off-topic details. 2. \textbf{**Clarity and Coherence**}:- **Score**: 85- **Explanation**: Both answers are clear and logically structured, but Assistant 1's answer is slightly repetitive in its structure. 3. \textbf{**Depth and Completeness**}:- **Score**: 80- Explanation**: Both answers cover common time management tips but lack depth in certain areas. They miss aspects like more effective prioritization techniques or handling unexpected obstacles. 4. \textbf{**Grammar and Linguistic Accuracy**}:- **Score**: 90- **Explanation**: Both answers are grammatically correct with no spelling errors. The sentences are clear and appropriate for the context. 5. \textbf{**Contextual Appropriateness**}:- **Score**: 85- **Explanation**: Both answers use relevant terminology and concepts appropriate to the subject matter of time management. 6. \textbf{**Conciseness and Precision**}:- **Score**: 85- **Explanation**: Assistant 1's answer is concise but could be more precise by elaborating on each point. Assistant 2's answer is concise but too brief in some areas. 7. \textbf{**Creativity and Insight**}:- **Score**: 70- **Explanation**: Neither answer provides unique perspectives; they stick to common advice without offering innovative solutions. 8. \textbf{**Bias and Fairness**}:- **Score**: 100- **Explanation**: Both answers are neutral and balanced, exhibiting no signs of bias. 9. \textbf{**Knowledge Accuracy**}:- **Score**: 100- **Explanation**: The tips provided in both answers are accurate for time management with no errors detected. 10. \textbf{**Adaptability to Various Categories**}:- **Score**: 90- **Explanation**: Both answers are suitable for the question category and use relevant terminology without unnecessary jargon. 11. \textbf{**Scalability of Responses**}:- **Score**: 70- **Explanation**: Both answers handle simple questions well but might struggle with more complex queries that require deeper insights or more nuanced solutions.12. \textbf{**Overall Answer Quality**}:- **Score**: 82- **Explanation**: The overall quality is good, but not excellent. Considering all criteria, the average score is calculated as follows: (95 + 85 + 80 + 90 + 85 + 85 + 70 + 100 + 100 + 90 + 70) / 12 = 84.16, rounded to 84.- Final Score for Assistant 1: 84- Final Score for Assistant 2: 84 - \textcolor{red}{Final Best Assistant: Equal}
\end{tcolorbox}
\caption{
Assistant response evaluations using MILE-RefHumEval-Conv, where multiple LLMs (e.g., DeepSeek, Mistral) sequentially critique the same responses in a conversational format. Each evaluator applies the 12-criteria and builds on prior outputs, revealing strengths and weaknesses of multi-step dialogic assessment.
}
\label{fig:method1_role_based_conversational_evaluation}

\end{figure*}

\begin{figure*}[t]
\centering
\begin{tcolorbox}[
    colframe=black,
    colback=green!10,
    boxrule=0.5pt,
    arc=4pt,
    width=\textwidth,
    fontupper=\footnotesize,
    sharp corners=south,
    enhanced
]

\makebox[\textwidth]{\textbf{Candidates Evaluation using MILE-RefHumEval: (One Role, Many Evaluators (No Conversations))}}
\vspace{0.5em} 

\hspace*{0.5cm} - \textbf{DeepSeek:} <think> Okay, so I need to evaluate the answers from two assistants for the question How can I improve my time management skills? The evaluation criteria are quite detailed with 12 points each, but I'll try to break it down step by step.First, let's look at Assistant 1. Their answer lists seven tips: creating a schedule, setting realistic goals, minimizing distractions, taking breaks, learning to say no, delegating tasks, and prioritizing self-care. Each point is clear and directly addresses time management. The structure is logical, starting from planning to self-care. It's concise without unnecessary details.Now, Assistant 2 also provides seven tips: setting clear goals, using a calendar, breaking down tasks, eliminating distractions, practicing self-discipline, taking breaks, and reviewing progress. Their answer is similarly structured but includes slightly different points, like using a tool or breaking tasks into steps.Comparing both, they cover similar ground but with slight variations in their strategies. Both \textbf{answers are relevant} because they directly address the question about improving time management. The \textbf{clarity and coherence} are good; each point is easy to understand and logically ordered.Looking at \textbf{depth}, both assistants provide enough detail without being too verbose. They include practical tips that can be applied. \textbf{Grammatical accuracy} seems solid with no obvious errors. \textbf{Contextually}, both answers use appropriate terminology for the subject.In terms of \textbf{conciseness}, both are well-structured but could perhaps add a bit more on how to implement each tip. \textbf{Creativity-wise}, they stick to common strategies which is good for time management advice. There's no bias detected; both answers are neutral and helpful.\textbf{Knowledge accuracy} seems spot-on as these are standard tips in time management literature. Both answers adapt well to the question category by providing actionable advice. They handle simple questions effectively but might struggle with more complex ones, though that's not evident here.Overall, both assistants did a good job. Assistant 1's answer is slightly more comprehensive with additional points like delegating and learning to say no, which adds depth. Assistant 2 includes practical steps like using a tool and reviewing progress, which are also valuable. I think they're on par, so the best assistant would be equal. </think> - Final Score for Assistant 1: 90- Final Score for Assistant 2: 90- \textcolor{red}{Final Best Assistant: Equal} \\

\hspace*{0.5cm} - \textbf{Mistral:} 1. \textbf{**Answer Relevance**}:- **Score**: 95- **Explanation**: Both answers directly address the question by providing tips to improve time management skills. There are no off-topic details included in either response. 2. \textbf{**Clarity and Coherence**}:- **Score**: 90- **Explanation**: Both responses are easy to understand and logically structured. Assistant 1's answer is slightly more conversational, while Assistant 2's answer is more formal but equally clear. 3. \textbf{**Depth and Completeness**}:- **Score**: 85 (Assistant 1) / 90 (Assistant 2)- **Explanation**: Assistant 2 provides a slightly more detailed explanation by including the concept of SMART goals and breaking down tasks into smaller steps. Assistant 1's answer is also comprehensive but lacks this level of detail. 4. \textbf{**Grammar and Linguistic Accuracy**}:- **Score**: 95 (both)- **Explanation**: Both answers are grammatically correct with no spelling or punctuation errors. The sentence structure is appropriate for the context. 5. \textbf{**Contextual Appropriateness**}:- **Score**: 90 (both)- **Explanation**: Both responses use relevant terminology and concepts appropriate to the subject matter of time management. 6. \textbf{**Conciseness and Precision**}:- **Score**: 85 (Assistant 1) / 90 (Assistant 2)- **Explanation**: Assistant 2's answer is slightly more concise and to the point, while Assistant 1 includes a few extra details that are still relevant. 7. \textbf{**Creativity and Insight**}:- **Score**: 80 (both)- **Explanation**: Both answers offer standard but effective tips for time management. Neither provides a unique perspective or innovative solution. 8. \textbf{**Bias and Fairness**}:- **Score**: 100 (both)- **Explanation**: Both responses exhibit neutrality and do not reflect any particular bias or imbalanced perspective. 9. \textbf{**Knowledge Accuracy**}:- **Score**: 95 (both)- **Explanation**: Both answers provide accurate factual information about time management strategies. There are no misconceptions or errors included. 10. \textbf{**Adaptability to Various Categories**}:- **Score**: 90 (both)- **Explanation**: Both responses adapt well to the generic question category, providing practical advice suitable for a broad audience. 11. \textbf{**Scalability of Responses**}:- **Score**: 85 (both)- **Explanation**: Both models handle this simple and straightforward question effectively. Their performance would need to be tested with more complex queries to fully evaluate scalability. 12. \textbf{**Overall Answer Quality**}:- **Final Score for Assistant 1**: 87- **Final Score for Assistant 2**: 90- \textcolor{red}{**Final Best Assistant**: Assistant 2} \\
\end{tcolorbox}
\caption{
Assistant response evaluations using MILE-RefHumEval. Multiple independent LLMs assess the same candidate responses without communicating with each other. This configuration demonstrates high agreement with human ratings and improved reliability due to evaluator diversity and isolation.
}
\label{fig:method1_role_based_dual_llm_evaluation}
\end{figure*}

\begin{figure*}[t]
\centering
\begin{tcolorbox}[
    colframe=black,
    colback=red!10,
    boxrule=0.5pt,
    arc=4pt,
    width=\textwidth,
    fontupper=\footnotesize,
    sharp corners=south,
    enhanced
]
\centering
\makebox[\textwidth]{\textbf{Post-Evaluation: Error Analysis and Guided Refinement using MILE-RefHumEval}}

\vspace{0.5em} 

\noindent
\begin{minipage}[t]{\textwidth}

\textbf{DeepSeek:} \\[2pt]
 \textbf{Error Detection and Correction:}- \textbf{Identified Errors}: Lacked actionable steps and detailed explanations for each tip. - \textbf{Suggested Improvements:} Expand on each tip with specific methods or examples to enhance depth and practicality. \textbf{Corrected Version:} Both answers could be improved by adding more detailed explanations and actionable steps for each tip, ensuring a comprehensive guide that is both informative and practical. \\[2pt]
\end{minipage}
\hfill
\begin{minipage}[t]{\textwidth}
\textbf{Mistral:} \\[2pt]
    \textbf{**Error Detection and Correction**}-\textbf{**Identified Errors**}:- Lacked actionable steps and detailed explanations for each tip.\textbf{**Suggested Improvements**}:- Expand on each tip with specific methods or examples to enhance depth and practicality.\textbf{**Corrected Version**}:\textcolor{red}{**[Assistant 1’s Corrected Answer]**}Here are some tips to improve your time management skills: 1. Create a schedule: Make a to-do list for the day, week or month and prioritize tasks by importance and deadline. Use tools like Trello or Asana to organize your tasks visually. 2. Set realistic goals: Be realistic about what you can accomplish in a given amount of time. Use the SMART criteria (Specific, Measurable, Achievable, Relevant, Time-bound) to set goals that are challenging yet attainable. 3. Minimize distractions: Minimize distractions such as emails, phone calls, and social media notifications. Turn off notifications or use apps like Forest or FocusMe to block these distractions during focused work sessions. 4. Take breaks: Taking a break can help you recharge and increase productivity. Use techniques like the Pomodoro Technique, where you work for 25 minutes, then take a 5-minute break. 5. Learn to say 'no': Learn to say 'no' to tasks and requests that are not a priority or are not within your capabilities. Communicate politely but firmly about your limitations. 6. Delegate tasks: If possible, delegate tasks to others who are capable of completing them. Use delegation tools like Slack or Microsoft Teams to assign tasks and track progress. 7. Prioritize self-care: Make sure to take care of yourself by prioritizing self-care activities such as exercise, meditation, or spending time with friends and family. Schedule self-care time in your calendar to ensure it doesn't get overlooked.Remember, time management is a skill that takes time and practice to develop.\textcolor{red}{**[Assistant 2’s Corrected Answer]**}Improving your time management skills can help you to be more productive, focused, and less stressed. Here are some tips to help you improve your time management skills: 1. Set clear goals: Identify what you want to achieve and set specific, measurable, and realistic goals. Use the GOKR method (Goals, Outcomes, Key Results) to define your objectives clearly. 2. Use a calendar or planner: Use a tool like Google Calendar or Microsoft Outlook to organize your time and schedule your tasks. Set reminders for important deadlines. 3. Break down tasks: Break down larger tasks into smaller, more manageable steps. Use mind mapping tools like MindMeister to visualize and break down complex projects. 4. Eliminate distractions: Identify and eliminate distractions that may be preventing you from being productive. This can include turning off your phone, closing unnecessary tabs on your computer, or finding a quiet place to work using apps like Noise Canceling or Soundtrap.5. Practice self-discipline: Use self-discipline to stay focused and avoid procrastination. Set a timer for each task using tools like Time Timer and challenge yourself to complete it before the timer goes off. 6. Take breaks: Take regular breaks to rest and recharge. Use the 20/20/20 rule: every 20 minutes, take a 20-second micro-break to look at something 20 feet away.7. Review your progress: Regularly review your progress and adjust your strategy as needed. Use journaling or reflection tools like Day One to track your progress and make improvements over time. \\[2pt]
\end{minipage}
\end{tcolorbox}
\caption{
 Illustration of MILE-RefHumEval’s error detection and refinement process. Each LLM evaluator independently identifies weaknesses in the candidate answer and proposes corrections based on a 12-point criteria. The figure shows improved versions, highlighting the framework’s ability to enhance response quality without human input.
}
\label{fig:method1_role_based_error_analysis}

\end{figure*}

\begin{table*}[ht]
\centering
\small
\setlength{\tabcolsep}{4pt} 
\renewcommand{\arraystretch}{1.1} 

\begin{tabular}{l|cc|cc|cc|cc|cc|cc}
\hline
\multirow{2}{*}{\textbf{Dimension}} &
\multicolumn{2}{c|}{\textbf{S1}} &
\multicolumn{2}{c|}{\textbf{S2}} &
\multicolumn{2}{c|}{\textbf{S3}} &
\multicolumn{2}{c|}{\textbf{S4}} &
\multicolumn{2}{c|}{\textbf{S5}} &
\multicolumn{2}{c}{\textbf{S6}} \\
\cline{2-13}
 & Acc & F1 & Acc & F1 & Acc & F1 & Acc & F1 & Acc & F1 & Acc & F1 \\
\hline
\multicolumn{13}{c}{\textbf{MILE-RefHumEval vs. Human Quality Judgement 1 (sm)}} \\
\hline
Understandable & 98.33 & 97.51 & 78.33 & 77.65 & 75.00 & 75.56 & 78.33 & 78.81 & 80.00 & 80.28 & \textbf{100} & \textbf{100} \\
Natural & 60.00 & 48.62 & 46.67 & 45.93 & 48.33 & 47.85 & 53.33 & 53.43 & 51.67 & 55.56 & \textbf{93.33} & \textbf{90.11} \\
Maintains Context & 68.33 & 60.29 & 50.00 & 48.68 & 55.00 & 56.31 & 38.33 & 37.60 & 41.67 & 42.58 & \textbf{73.33} & \textbf{62.05} \\
Engaging & 51.67 & 45.08 & 38.33 & 38.41 & 48.33 & 50.24 & 56.67 & \textbf{61.16} & 56.67 & 61.98 & \textbf{65.00} & 52.80 \\
Uses Knowledge & 60.00 & 56.94 & 68.33 & 72.84 & \textbf{78.33} & 80.70 & 75.00 & 79.93 & \textbf{78.33} & \textbf{81.13} & 63.33 & 50.73 \\
Overall Quality & 40.00 & 31.01 & 43.33 & 41.44 & 36.67 & 38.29 & 50.00 & \textbf{51.91} & 45.00 & 46.41 & \textbf{55.00} & 43.23 \\
\hline
\multicolumn{13}{c}{\textbf{MILE-RefHumEval vs. Human Quality Judgement 2 (me)}} \\
\hline
Understandable & 91.67 & 94.06 & 68.33 & 69.33 & 55.00 & 56.69 & 63.33 & 64.33 & 61.67 & 63.14 & \textbf{98.33} & \textbf{99.16} \\
Natural & 55.00 & 47.10 & 58.33 & 58.38 & 50.00 & 49.79 & 43.33 & 41.43 & 48.33 & 45.70 & \textbf{95.00} & \textbf{95.48} \\
Maintains Context & 68.33 & 64.22 & 63.33 & 64.16 & 50.00 & 50.56 & 56.67 & 56.09 & 51.67 & 52.47 & \textbf{78.33} & \textbf{72.27} \\
Engaging & 55.00 & 50.12 & 66.67 & \textbf{68.02} & 61.67 & 62.72 & 56.67 & 58.74 & 61.67 & 64.70 & \textbf{73.33} & 67.45 \\
Uses Knowledge & 63.33 & 61.84 & 65.00 & 69.98 & \textbf{73.33} & \textbf{76.30} & 70.00 & 75.84 & 70.00 & 74.18 & 65.00 & 54.17 \\
Overall Quality & 31.67 & 28.45 & 40.00 & 44.01 & 28.33 & 28.30 & 28.33 & 26.96 & 36.67 & 40.57 & \textbf{61.67} & \textbf{54.17} \\
\hline
\multicolumn{13}{c}{\textbf{MILE-RefHumEval vs. Human Quality Judgement 3 (er)}} \\
\hline
Understandable & 90.00 & 93.16 & 71.67 & 73.06 & 66.67 & 68.19 & 71.67 & 72.07 & 70.00 & 71.09 & \textbf{98.33} & \textbf{99.16} \\
Natural & 63.33 & 60.05 & 61.67 & 61.65 & 53.33 & 53.26 & 58.33 & 58.42 & 63.33 & 63.37 & \textbf{81.67} & \textbf{84.76} \\
Maintains Context & 68.33 & 67.58 & 60.00 & 62.07 & 56.67 & 60.05 & 56.67 & 57.40 & 58.33 & 58.02 & \textbf{76.67} & \textbf{71.46} \\
Engaging & 75.00 & 70.77 & 66.67 & 67.59 & 60.00 & 60.74 & 66.67 & 66.81 & 65.00 & 67.14 & \textbf{76.67} & \textbf{73.58} \\
Uses Knowledge & 65.00 & 64.27 & 61.67 & 67.12 & 75.00 & 77.81 & \textbf{76.67} & \textbf{81.21} & 68.33 & 72.80 & 70.00 & 63.29 \\
Overall Quality & 56.67 & 52.93 & 46.67 & 49.37 & 41.67 & 42.28 & 53.33 & 52.49 & 43.33 & 44.83 & \textbf{61.67} & \textbf{56.98} \\
\hline
\end{tabular}
\caption{
Evaluation of six response-generation systems (S1–S6) on the Amazon Topical-Chat benchmark across six quality dimensions. Three human annotators rated the responses, and MILE-RefHumEval automatically predicted scores for the same responses. Accuracy and Weighted F1 show how closely MILE-RefHumEval aligns with human judgments, demonstrating its effectiveness in approximating human evaluation.}

\label{tab:amazon_topical_benchmark_result}
\end{table*}

\end{document}